\pdfoutput=1
\documentclass[11pt]{article}

\usepackage[letterpaper,margin=1in]{geometry}
\usepackage[utf8]{inputenc}
\usepackage[T1]{fontenc}
\usepackage{times}
\usepackage[numbers,sort]{natbib}
\usepackage{hyperref}
\usepackage{url}
\usepackage{booktabs}
\usepackage{amsfonts}
\usepackage{microtype}
\usepackage{graphicx}
\usepackage{amsmath}
\usepackage{amssymb}
\usepackage{subcaption}
\usepackage{enumitem}

\newcommand{\method}{\textsc{UniVL}}

\title{UniVL: Unified Vision-Language Embedding \\for Spatially Grounded Contextual Image Generation}

\author{
  Jiayun Wang \quad Yu Wang \quad Weijie Gan \quad Zhenting Wang \quad Wei Wei \\
  Center for Advanced AI, Accenture
}

\date{}

\begin{document}

\maketitle

\vspace{-1em}
\begin{abstract}

We introduce \emph{spatially grounded contextual image generation}, a new controllable image generation task, and propose \method{} (Unified Vision-Language image generation) for it. Instead of supplying a reference image and a global text prompt through two separate encoders (vision and language), \method{} binds semantics to spatial locations directly from a \emph{single unified visual input} where the textual instruction is rendered onto the spatial mask, eliminating the standalone text encoder at inference while following the user's per-region \emph{what-goes-where} intent.
For the task, we propose a framework in which the \method{} encoder---adapted from an optical-character-recognition-pretrained backbone---reads the unified condition optically, producing a \method{} embedding $f_{\text{VL}}$ that fuses visual and semantic intents to spatial locations, packed as a single token sequence. A two-stage pipeline aligns \method{} in VAE embedding space and then conditions a pretrained diffusion backbone entirely on \method{} embeddings, eliminating the standalone text encoder (e.g., T5).
The reframing is deliberately minimalist for text, but the empirical payoff is large. On \method-ImgGen, a benchmark of $477$K mask-annotated images that we construct to support training and evaluation, \method{} achieves \emph{superior image quality} over text-prompted baselines (FID: $14 \to 11$, PSNR: $16 \to 20$) while \emph{eliminating the text encoder entirely}, reducing inference TFLOPs by up to 52\% and runtime by up to 44\%. 
Additional ablation studies verify components of different parts of the proposed method, paving way for efficient and spatially grounded image generation with unified conditioning paradigm.
\end{abstract}

\section{Introduction}

\begin{figure}[t]
\vspace{-1em}
    \centering
    \includegraphics[width=\linewidth]{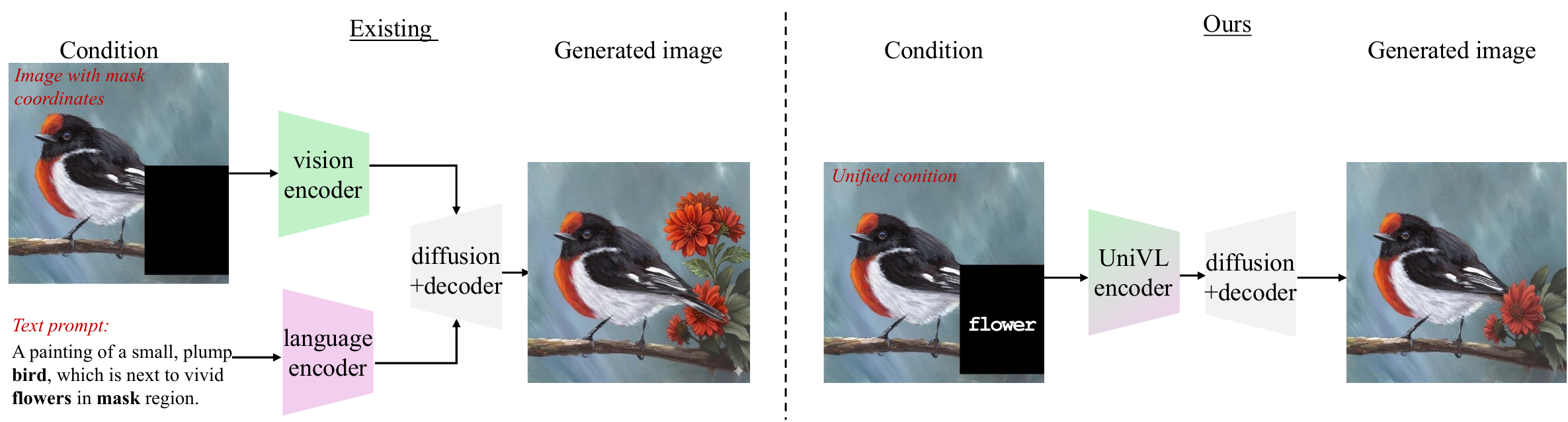}
    \caption{Conditioning paradigm for contextual image generation: existing (\emph{left}) vs.\ our \method{} (\emph{right}). The user's intent is the same---inpaint a target region with a desired object. Existing methods require two inputs through two separate domains: a masked image (vision) plus a global text prompt (language), each processed by its own encoder before the diffusion and decoder. In \method, the same intent is captured by a \emph{single unified visual condition}: our pipeline automatically renders the text label as pixels inside the mask. The \method{} encoder (an OCR-pretrained vision backbone adapted to our task) then reads spatial layout and textual semantics in one pass. The text encoder is eliminated, yet the generated image preserves the same semantic intent (``flower'') at specified spatial locations.}
    \label{fig:teaser}
    \vspace{-0.5em}
\end{figure}

Text-to-image diffusion models can generate high-fidelity images from natural language prompts~\citep{tan2025omni,dhariwal2021diffusion}, and recent controllable generation methods add spatial signals such as depth maps or bounding boxes to improve user control~\citep{zhang2023controlnet,li2023gligen}. Yet a fundamental issue remains: \emph{text prompts} are global and carry rich semantics but no spatial grounding, while \emph{structural controls} carry local and precise geometry but no semantic content. Bridging this gap---specifying \emph{what} should appear \emph{where} in a single unified signal---is still an open problem.

Such a unified signal has, in fact, already emerged in visual \emph{understanding} tasks such as document parsing and visual question answering. Recent work on optical context compression~\citep{deepseekocr,deepseekocr2} renders text as pixels and processes it with a vision encoder, producing representations that compress thousands of language tokens into fewer than 256 vision tokens while preserving fine-grained spatial structure. The benefit of this approach is conceptually simple: spatial constraints in document images become semantic constraints at pixel locations. This is precisely the property that controllable generation currently lacks. We therefore ask: can optical compression be reoriented to a unified conditioning signal for controllable generation?

\begin{figure}[t]
\vspace{-1em}
    \centering
    \includegraphics[width=0.94\linewidth]{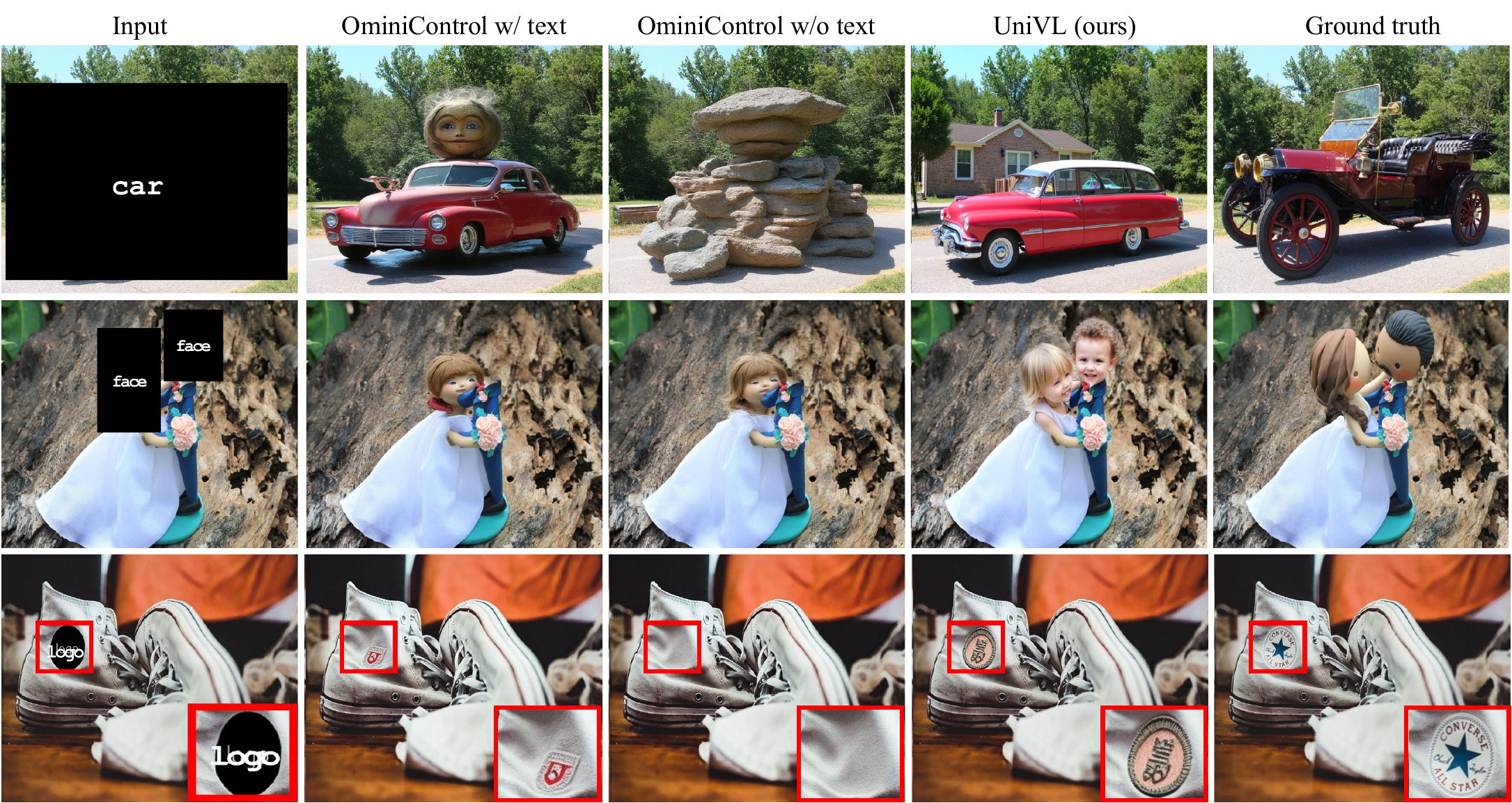}
    \caption{Qualitative comparison to OminiControl baseline \citep{omini}. ``OminiControl w/ text'' is only method with text encoder and additional overhead. 
    \emph{Row 1} (single mask, ``car''): OminiControl with prompt hallucinates a doll over a car, the no-prompt variant collapses into a rock formation, while \method{} produces a clean prompt-faithful car. \emph{Row 3} (zero-shot test on free-form mask): \method{} respects the user-provided mask boundary and generates content that conforms to its shape, whereas the baselines either ignore the irregular boundary or default to bounding-box-shaped fills (details in Appendix~\ref{app:freeform}). Input is for \method, see Fig.\ref{fig:teaser} for details.}
    \vspace{-1em}
    \label{fig:qualitative}
\end{figure}

We answer it with a new task called \emph{spatially grounded contextual image generation}. 
The user provides a reference image, draws one or more mask regions, and supplies a short text label for each region. Our pipeline automatically renders each label as pixels on its mask, producing a single \emph{contextual condition} image. The model must then generate content matching each rendered text on its mask while preserving the rest of visuals. This formulation has two properties that motivate our approach. First, it \emph{co-locates semantics and spatial position in one input}: the text label and the visual context occupy the same image at the same coordinates, so the user can specify what should appear in each region with explicit per-region grounding, and the model can be trained to handle multi-region edits in a single forward pass (Fig.\ref{fig:qualitative}). Second, it \emph{eliminates the standalone text encoder}: a single OCR-pretrained vision encoder handles both the visual context and the textual instruction, removing the heavy text encoder like T5 \cite{roberts2022t5x} and yielding substantial efficiency gains.

We instantiate this task with \method{} (Unified Vision-Language image generation), a framework built on an inpainting-capable diffusion backbone (FLUX.1-dev~\citep{greenberg2025demystifying}). \method{} adds two training stages: (1)~\emph{\method{} embedding alignment}, which fine-tunes the \method{} encoder (adapted from DeepSeek-OCR's OCR-pretrained encoder \cite{deepseekocr}) so its output $f_{\text{VL}}$ reconstructs VAE latents of the target image (Fig.\ref{fig:training_scheme}, top) and allows direct uses of VAE decoder; and (2)~\emph{diffusion fine-tuning}, which conditions the diffusion model entirely on \method{} embeddings (Fig.\ref{fig:training_scheme}, bottom). To support training and evaluation, we construct the \emph{\method-ImgGen} benchmark of ${\sim}477$K mask-annotated images (${\sim}710$K mask-text records). Without any text encoder, \method{} improves spatial-semantic binding (FID: $13.5 \to 11.1$ over OminiControl) while cutting inference TFLOPs by up to 52\% and runtime by up to 44\%. An optional \emph{\method+}text encoder hybrid further closes the small remaining alignment gap, indicating that our reframing is compatible with---and not a replacement for---the standard text+image conditioning paradigm when an extra text encoder is acceptable.

We summarize our contributions as follows: (1)~\textbf{A new controllable image generation task with a unified condition}: we formalize \emph{spatially grounded contextual image generation}, which reframes the conditioning interface from two separate domains (vision + language) into a single unified visual input where text is rendered onto the spatial mask. We construct the \textit{\method-ImgGen benchmark} ($477$K training images with mask-text annotations) as a reusable resource. (2)~\textbf{\method: Unified Vision-Language image generation framework}: we show that OCR-pretrained optical embeddings can be \emph{adapted} into effective conditioning signals for diffusion through a two-stage alignment-and-finetune pipeline plus a small set of training-time design choices that we ablate (Section~\ref{sec:ablations}). (3)~\textbf{Significant efficiency without quality degradation}: by eliminating the text encoder, \method{} reduces token size by up to 98\%, encoder parameters by $92\%$, TFLOPs by up to 52\%, and runtime by up to 44\% (Table \ref{tab:efficiency}), while \emph{matching or exceeding} prompted baselines on image quality.

\section{Related Work}

\textbf{Diffusion-based image generation and editing.}
Diffusion models~\citep{ho2020ddpm,dhariwal2021diffusion} and their latent variants~\citep{rombach2022ldm} are the dominant text-to-image foundation; region-aware editing has been built on top through inpainting and instruction-tuned variants~\citep{lugmayr2022repaint,hertz2022prompt,brooks2023instructpix2pix}. \emph{Our difference}: rather than supply a separate text prompt alongside a mask, we render the textual instruction directly inside the mask and remove the standalone text encoder.

\textbf{Unified vision-language representation learning.}
Large-scale VL pretraining yields unified image-text embeddings that transfer broadly across tasks~\citep{radford2021clip,li2023blip2}. A more recent line targets a single backbone that decodes both language and images, including Janus-series~\citep{wu2024janus,chen2025januspro}, Ming-Omni~\citep{inclusionai2025mingomni} and Omni-Video~\citep{tan2025omni}. The closest work to ours is DeepSeek-OCR~\citep{deepseekocr}, which compresses text into a unified vision-language representation. \emph{Our difference}: CLIP-style representations are not driven through to generation; Janus-style unified models retain \emph{separate} encoders for the input modalities and unify only the decoder; DeepSeek-OCR targets LLM/VLM understanding rather than image generation. \method{} instead uses a unified \emph{encoder} as the conditioning interface for image synthesis.

\textbf{Spatially grounded controllable generation.}
Before diffusion, GANs explored spatial control via image-to-image translation~\citep{zhu2017cyclegan}, semantic-layout-to-image generation~\citep{park2019spade}, and high-resolution synthesis~\citep{karras2019stylegan}, establishing the principle of using auxiliary spatial signals to constrain \emph{where} content appears. Diffusion models extended these ideas: ControlNet adds structured spatial signals such as edges and depth~\citep{zhang2023controlnet}, GLIGEN grounds generation with layout boxes and class labels~\citep{li2023gligen}, and inpainting-style methods (BrushNet~\citep{ju2024brushnet}, PowerPaint~\citep{zhuang2024powerpaint}, FLUX.1 Fill) tackle masked region completion. Subject-driven variants such as AnyDoor~\citep{chen2024anydoor} and Paint-by-Example~\citep{yang2023paintbyexample} condition on a reference image, and OmniGen~\citep{xiao2024omnigen} unifies multiple generative tasks behind one transformer while still retaining a text encoder. \emph{Our difference}: to our knowledge, \method{} is the first to use a \emph{unified vision encoder} as the conditioning interface for diffusion-based controllable generation, encoding both spatial layout and textual semantics from a single visual input.

\textbf{Glyph-aware diffusion generation.}
A complementary line tackles \emph{rendering} legible text inside generated images: AnyText~\citep{tuo2023anytext} introduces an auxiliary text-control branch on top of latent diffusion to synthesize multilingual glyphs at user-specified positions; GlyphDraw~\citep{ma2023glyphdraw} biases attention via glyph-image embeddings; TextDiffuser~\citep{chen2023textdiffuser} adds a layout transformer that predicts character bounding boxes and conditions on character masks. \emph{Our difference}: these methods treat the rendered text as the \emph{output target}---the goal is to produce an image with that text legibly drawn on top; whereas \method{} treats rendered text as a \emph{conditioning interface}, optically compressing a textual instruction into pixels that are read by a vision encoder to drive non-textual content generation inside the masked region.

\section{The \method-ImgGen Benchmark}
\label{sec:benchmark}

To support training and evaluation of spatially grounded contextual image generation, we construct \method-ImgGen, a benchmark comprising $477$K unique images and $710$K mask-text records. We first define the task, then describe how the benchmark is built and used for training and evaluation.

\textbf{Task definition.} We introduce \emph{spatially grounded contextual image generation}, an inpainting-style controllable task with a unified visual interface. The user provides a source image $X \in \mathbb{R}^{H \times W \times 3}$, draws one or more mask regions $\{m_i\}_{i=1}^{K}$, and supplies a text label $\ell_i$ per region. Our pipeline produces a single \emph{contextual condition} $C_I$ by masking each region (e.g., filling with black) and rendering its label as pixels inside the mask. The model must then generate an output $\hat{X}$ that (a)~matches $\ell_i$ within each $m_i$ and (b)~preserves the source content outside all masks. Existing paradigms split this signal across two streams---global text prompts~\citep{rombach2022ldm,ramesh2022dalle2}, layout boxes with labels~\citep{li2023gligen}, or structural cues such as edges/depth~\citep{zhang2023controlnet}---each weakly coupling location and semantics. Our task instead unifies both into a single visual input that the \method{} encoder reads jointly, naturally supporting multi-region composability ($K \geq 1$) in one forward pass.

\textbf{Benchmark overview.} \method-ImgGen comprises 710K mask-text records spanning $477$K unique images and $28$K unique class-name phrases, drawn from four sources: a \emph{mask} category built from a 238K-image LAION-5B subset~\citep{schuhmann2022laion} via Grounding DINO~\citep{liu2023grounding} detection plus size and CLIP-similarity filtering, and three image-editing categories (\emph{add}, \emph{replace}, \emph{extract}) following ImgEdit~\citep{ye2025imgedit}. The mask category provides multibox/spatial-localization training; the editing categories provide same-image-different-text pairs that prevent the model from collapsing to image-reconstruction shortcuts. While the benchmark uses rectangular masks for simplicity, our method generalizes to free-form masks (Appendix~\ref{app:freeform}).  
Full construction details, augmentation, and class-name statistics are in Appendix~\ref{app:benchmark}. We hold out 3{,}000 samples for evaluation (1{,}500 single-mask and 1{,}500 multi-mask with $N{\in}[1,5]$); the training/eval split is described in Appendix~\ref{app:benchmark}.

\begin{figure}[t]
    \centering
    \vspace{-1em}
    \includegraphics[width=0.95\linewidth]{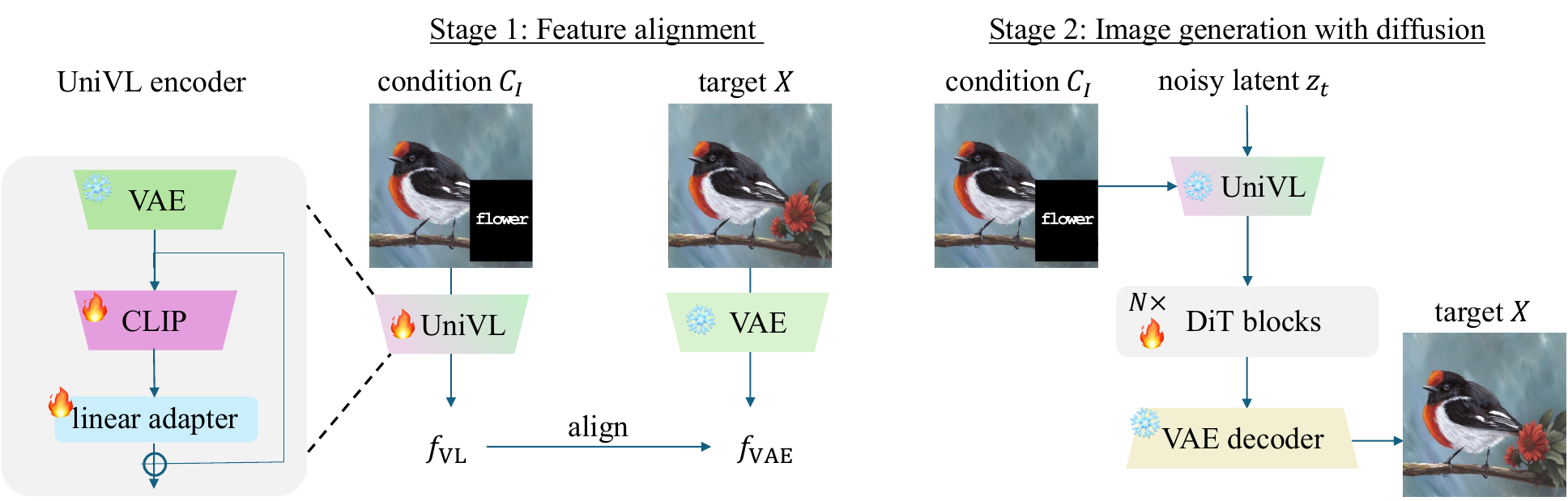}
    \caption{\method{} training pipeline. The \method{} encoder (left, pretrained on OCR tasks) consists of a frozen VAE encoder, a trainable CLIP backbone and a trainable linear adapter; the VAE features bypass CLIP via a skip connection and are mask-aware-fused with the CLIP features (Eq.~\ref{eq:mask_aware_add}). \textbf{Stage~1 (Feature alignment):} the \method{} encoder takes the contextual condition $C_I$ and is trained so its output $f_{\text{VL}}$ reconstructs the VAE latent $z_0 = \mathcal{E}(X)$ of the target image $X$. \textbf{Stage~2 (Image generation with diffusion):} the \method{} encoder produces $f_{\text{VL}}$ from $C_I$ and feeds it (alongside the noisy latent $z_t$) into the DiT; the VAE decoder then yields the target $X$. }
    \label{fig:training_scheme}
\end{figure}
\section{\method: Unified Vision-Language Embedding for Contextual Generation}
\label{sec:method}

\subsection{Architecture overview}
\label{sec:univl_overview}

We build \method{} on top of an inpainting-capable diffusion backbone (DiT-based). Inpainting pretraining provides a useful prior for spatially localized generation (\emph{where} to generate) before we introduce the semantic signal (\emph{what} to generate) via \method{} conditioning. Rather than train this stage ourselves, we use a pretrained diffusion model \citep{greenberg2025demystifying} as the starting point and focus on the \method{} encoder and the two training stages that produce and exploit the \method{} embedding (Fig.\ref{fig:training_scheme}).

\textbf{Diffusion transformers (DiT).} Diffusion Transformers~\citep{peebles2023scalable} replace the U-Net backbone of latent diffusion models with a transformer operating on patchified latent tokens. Given a latent representation $z_0 = \mathcal{E}(X)$ from a VAE encoder $\mathcal{E}$, the forward process adds Gaussian noise to produce $z_t$ at timestep $t$, and a transformer $\epsilon_\theta$ is trained to predict the noise via the standard denoising objective $\mathcal{L}_{\text{diff}} = \mathbb{E}_{z_0, \epsilon, t}\big[\|\epsilon - \epsilon_\theta(z_t, t, c)\|^2\big]$, where $c$ denotes the conditioning signal. At inference, iterative denoising from $z_T \sim \mathcal{N}(0, I)$ followed by VAE decoding produces the output image. Recent methods inject spatial control into DiT by concatenating condition tokens with the noisy latent sequence; OminiControl~\citep{omini} proposes a unified token-sequence strategy that interleaves condition and image tokens within the transformer. Building upon the DiT architecture with FLUX.1~\citep{greenberg2025demystifying}, we aim to develop a minimal and controllable generation framework that accepts flexible control signals.

\textbf{From two conditioning encoders to one.} In standard conditional diffusion frameworks \cite{omini,greenberg2025demystifying}, the DiT consumes three inputs: an image-conditioning latent $C_I$ (the VAE encoding of a reference or masked image), a text-prompt embedding $C_T$ (from a separate text encoder such as T5), and the initial noise $z_t \sim \mathcal{N}(0, I)$. The DiT iteratively denoises $z_t$ conditioned on the image and text signals; the resulting clean latent $z_0$ is decoded by the VAE to produce the target image $X$. \method{} changes the conditioning interface: it \emph{eliminates} $C_T$ and lets the contextual condition $C_I$ (the user-provided RGB image with text rendered into the mask, defined in Section~\ref{sec:benchmark}) carry both the visual context and the textual instruction. The \method{} encoder (Section~\ref{sec:univl_encoder}) maps $C_I$ to a unified embedding $f_{\text{VL}}$ that the DiT consumes alongside $z_t$. Removing the text encoder branch yields the efficiency gains reported in Section~\ref{sec:efficiency} while keeping the same DiT architecture and training objective.

\subsection{\method{} encoder architecture}
\label{sec:univl_encoder}

The \method{} encoder (green box in Fig.\ref{fig:training_scheme}), adapted from DeepSeek-OCR's OCR encoder~\citep{deepseekocr}, takes the contextual condition $C_I$ and produces a token sequence $f_{\text{VL}}$ that conditions the DiT. It has three components: a \emph{frozen VAE encoder} that extracts low-level visual features $f_{\text{VAE}}$, a \emph{trainable CLIP backbone} (initialized from DeepSeek-OCR's OCR-pretrained weights) that extracts high-level semantic features $f_{\text{CLIP}}$, and a small \emph{trainable linear adapter} that maps token dimensions to match the DiT's conditioning interface. We follow DeepSeek-OCR's recipe of pairing a low-level visual encoder with a high-level semantic encoder, but replace SAM with the FLUX VAE for two reasons: (i)~the VAE plays the same role as SAM in this stack---encoding low-level visual content of the contextual condition---and (ii)~the VAE is the same encoder used by the diffusion backbone, so $f_{\text{VAE}}$ is already in the latent space the DiT consumes, simplifying downstream alignment. The frozen VAE decoder can also be directly used for decoding latents. CLIP is retained for its high-level semantic features (especially after OCR pretraining, which adapts CLIP to read rendered text). Because the encoder is initialized from a publicly released OCR-pretrained checkpoint trained on a large and diverse corpus ($30$M PDF pages, $20$M scene-OCR images, plus chart/formula/geometry parsing data; Appendix~\ref{app:ocr_choice}), we do not re-run OCR pretraining---we adapt this encoder for our generation conditioning task in Stage~1 below. Note that the \method{} encoder does not perform explicit character recognition; rather, OCR pretraining endows it with a learned mapping from \emph{text-as-pixels patterns} to semantic features, which Stage~1 then aligns with the VAE latent space. 

\textbf{Mask-aware fusion.} The two feature streams are combined by a mask-aware additive operator. Let $M \in \{0,1\}^{H \times W}$ be the binary mask aggregating all user-specified regions ($M_{ij}=1$ inside any masked region, $0$ outside). The \method{} encoder output is
\begin{equation}
    f_{\text{VL}} \;=\; f_{\text{VAE}} \odot (1 - M) \;+\; f_{\text{CLIP}} \odot M,
    \label{eq:mask_aware_add}
\end{equation}
i.e., outside-mask positions are dominated by VAE features (preserving background context) and inside-mask positions are dominated by CLIP features (carrying the semantic content read from the rendered text). The trainable adapter is then applied on top of $f_{\text{VL}}$.

\subsection{Two-stage training}
\label{sec:univl_training}

\textbf{Stage 1: \method{} embedding alignment.}
We adapt the \method{} encoder to produce a \method{} embedding $f_{\text{VL}}$ suitable for conditioning diffusion generation. As shown in Fig.\ref{fig:training_scheme}~(top), the \method{} encoder processes the contextual condition $C_I$ and produces $f_{\text{VL}}$, while the frozen VAE $\mathcal{E}$ processes the target image $X$ and produces the latent $z_0 = \mathcal{E}(X)$. We minimize the feature alignment objective
\begin{equation}
    \mathcal{L}_{\text{align}} = \|f_{\text{VL}} - z_0\|_2^2,
    \label{eq:align}
\end{equation}
aligning the \method{} embedding of the text-on-mask input with the visual representation of the desired output. The VAE encoder is fully frozen; the CLIP backbone is updated through a low-rank LoRA adapter, and the linear adapter on top is trained from scratch in full precision. This combination keeps trainable parameters small (the OCR-pretrained CLIP weights remain largely intact) while still letting the encoder adapt to the natural-image distribution and rendered-text inputs.

\textbf{Stage 2: Diffusion fine-tuning with \method{} conditioning.}
The \method{} encoder (initialized from Stage~1) processes the contextual condition $C_I$ to produce the \method{} embedding $f_{\text{VL}}$, which replaces the conventional text embedding as the conditioning signal for the DiT. As shown in Fig.\ref{fig:training_scheme}~(bottom), $f_{\text{VL}}$ is injected into the DiT blocks alongside the noisy latent $z_t$, and the model is trained with the standard denoising objective $\mathcal{L}_{\text{diff}} = \mathbb{E}_{z_0, \epsilon, t}\!\left[\|\epsilon - \epsilon_\theta(z_t, t, f_{\text{VL}})\|^2\right]$, augmented with the auxiliary feature-reconstruction and CLIP terms from Stage~1:
\begin{equation}
\label{eq:full}
    \mathcal{L}_{\text{stage-2}} = \mathcal{L}_{\text{diff}} +  \mathcal{L}_{\text{align}} + \lambda_{\text{clip-img}}\, \mathcal{L}_{\text{clip-img}} + \lambda_{\text{clip-txt}}\, \mathcal{L}_{\text{clip-txt}}.
\end{equation}
In addition to the feature alignment objective $\mathcal{L}_{\text{align}}$, we introduce two auxiliary CLIP-based losses that play a critical role in Stage~2: a CLIP image loss $\mathcal{L}_{\text{clip-img}}$ that measures the $\ell_2$ distance between the \method{} encoder output $f_{\text{VL}}$ and the CLIP vision embedding corresponding to patches within the mask; and a CLIP text loss $\mathcal{L}_{\text{clip-txt}}$ that measures the $\ell_2$ distance between the pooled masked condition and the CLIP text embedding of the class name. 
Keeping the feature alignment, CLIP image, and CLIP text losses active during diffusion training prevents $f_{\text{VL}}$ from drifting away from the visual-semantic space established in Stage~1. In Stage~2 we update the DiT through a LoRA adapter ($r{=}4$, $\alpha{=}4$) and continue to update the \method{} encoder's CLIP-LoRA and linear adapter (both initialized from Stage~1); the VAE encoder is fully frozen. The training uses condition dropout, which independently drops the entire condition image, the VAE-stream tokens, or the CLIP-stream tokens to enable classifier-free guidance at inference (details in Appendix~\ref{app:dropout}).

 The two-stage pipeline introduces additional \emph{training} overhead relative to end-to-end joint fine-tuning, but yields measurably higher quality (Table~\ref{tab:ablation_components}, ``w/o feature alignment''). The Stage 1 + Stage 2 wall-clock is reported in Section~\ref{sec:exp_setup}; the efficiency claims in Table~\ref{tab:efficiency} concern \emph{inference} only, where the \method{} encoder is a single forward pass and Stage~1 contributes no overhead. The training overhead is therefore a one-time cost amortized across all downstream inferences.

\section{Experimental Results}

We evaluate \method{} along four axes: (i) generation quality and alignment against controllable-generation baselines on the \method-ImgGen benchmark (Section~\ref{sec:main_results}), (ii) qualitative behavior under multi-region conditioning (Section~\ref{sec:multiregion}), (iii) computational efficiency (Section~\ref{sec:efficiency}), and (iv) a series of ablations isolating the contribution of each component (Section~\ref{sec:ablations}).

\subsection{Experimental Setup}
\label{sec:exp_setup}

\textbf{Training, evaluation and metrics.}
We train on the \method-ImgGen benchmark (Section~\ref{sec:benchmark}), which contains $710$K mask-text records spanning four sources: a mask category derived from a LAION-5B~\citep{schuhmann2022laion} via Grounding DINO~\citep{liu2023grounding} detection plus filtering, and three image-editing categories (add, replace, extract) following ImgEdit~\citep{ye2025imgedit}.
The evaluation set consisting of 3{,}000 samples evenly split between 1{,}500 single-mask samples (375 each from the four source categories: mask, object addition, replacement, and extraction) and 1{,}500 multi-mask samples ($N{\in}[1,5]$ masks per image). See Appendix~\ref{app:benchmark} for details.
We report metrics in two groups. For {image quality}, we use FID (distribution-level, lower is better), SSIM, PSNR and MUSIQ~\citep{ke2021musiq} (higher is better). For {alignment}, we report CLIP$_{\text{txt}}$ and CLIP$_{\text{img}}$ (global text-image and image-image similarity) along with their region-level counterparts region-CLIP$_{\text{txt}}$ and region-CLIP$_{\text{img}}$---CLIP similarity between the cropped generated region and the text label or the ground-truth crop, respectively (higher is better).

\textbf{Implementation.}
All experiments use FLUX.1-dev~\citep{greenberg2025demystifying} as the base diffusion model, trained with AdamW in bfloat16: Stage~1 (alignment) takes ${\approx}12.6$ hours over 30K steps on 2$\times$H100 80GB GPUs, and Stage~2 (diffusion fine-tuning, LoRA $r{=}4$) takes ${\approx}97.3$ hours over 40K steps on 6$\times$H100 80GB GPUs. Auxiliary CLIP-image and CLIP-text losses ($\lambda_{\text{clip-img}}{=}1.0$, $\lambda_{\text{clip-txt}}{=}0.8$) are applied during Stage~2; their importance is validated in Appendix~\ref{app:ablation_loss}. At inference we use 28 denoising steps with guidance scale 1.0 (no classifier-free guidance, since no text prompt is used). Full per-stage hyperparameters and loss details are in Appendix~\ref{app:hparams} and Appendix~\ref{app:loss}.

\textbf{Baselines.}
Our primary baseline is {OminiControl}~\citep{omini}, a state-of-the-art controllable generation method for FLUX that uses token-sequence conditioning for spatial and subject-driven control. We evaluate it in two settings (both finetuned on the same training data as \method{} for fairness): (1)~with its standard T5 text prompt (reference only), and (2)~without text prompt, which provides a fair comparison at identical computational cost and input modality to \method. Neither \method{} nor OminiControl w/o prompt uses a text encoder. FLUX.1 Fill is also finetuned on the same data.

\subsection{Main Results}
\label{sec:main_results}

Table~\ref{tab:main_results} compares \method{} against OminiControl~\citep{omini}, a state-of-the-art controllable generation method for FLUX. We evaluate OminiControl in two settings: with the original T5 text prompt (its standard configuration), and without text prompt (matching our setting where no linguistic text input is provided). The no-prompt variant is the \emph{fair} comparison: both methods receive identical spatial conditioning and use the same computational budget, with neither having access to a text encoder.

\begin{table}[t]
    \centering
    \vspace{-1em}
    \caption{Comparison on the \method-ImgGen benchmark (single and multiple masks), split by whether the method uses a text encoder. The \emph{w/o text encoder} setting (top) is our primary focus: it provides a fair comparison at identical computational cost and input modality to \method. The \emph{w/ text encoder} setting (bottom) reports prompted methods for reference. Reported numbers are the mean across 5 inference seeds; best per metric is in \textbf{bold}, second-best \underline{underlined}.  }
    \label{tab:main_results}
    \resizebox{\textwidth}{!}{
    \begin{tabular}{l cccc c cccc}
        \toprule
        & \multicolumn{4}{c}{Image Quality} & & \multicolumn{4}{c}{Alignment} \\
        \cmidrule(lr){2-5} \cmidrule(lr){7-10}
        Method & FID$\downarrow$ & SSIM$\uparrow$ & PSNR$\uparrow$ & MUSIQ$\uparrow$ & & CLIP$_{\text{txt}}\uparrow$ & CLIP$_{\text{img}}\uparrow$ & region-CLIP$_{\text{txt}}\uparrow$ & region-CLIP$_{\text{img}}\uparrow$ \\
        \midrule
        \multicolumn{10}{l}{\emph{Without text encoder (primary focus)}} \\
        ControlNet-FLUX                  & \underline{13.98} & \underline{0.734} & \underline{18.22} & 70.50             & & \underline{0.250} & \underline{0.856} & 0.249             & 0.689             \\
        OminiControl~\citep{omini}       & 20.50             & 0.719             & 16.45             & \underline{72.05} & & 0.249             & 0.849             & \underline{0.271} & \underline{0.742} \\
        \textbf{\method{} (ours)}          & \textbf{11.13}    & \textbf{0.754}    & \textbf{19.61}    & \textbf{73.36}    & & \textbf{0.262}    & \textbf{0.886}    & \textbf{0.279}    & \textbf{0.768}    \\
        \midrule
        \multicolumn{10}{l}{\emph{With text encoder (reference)}} \\
        BrushNet~\citep{ju2024brushnet} & 18.33 & 0.611 & 17.38 & 73.21 & & 0.272 & 0.836 & 0.247 & 0.757\\
         PowerPaint~\citep{zhuang2024powerpaint} & 29.4 & 0.519 & 14.17 & 63.6 && 0.266 & 0.766 & 0.256 & 0.699 \\
        ControlNet~\citep{zhang2023controlnet}      & 23.93             & 0.684             & 15.30             & 69.82             & & 0.264             & 0.859             & 0.282             & 0.763             \\
        OminiControl~\citep{omini}                  & 13.50             & 0.718             & 16.13             & \underline{72.81} & & 0.270             & 0.871             & \textbf{0.298}    & \underline{0.770} \\
        ControlNet-FLUX                             & 13.77             & 0.731             & \underline{18.02} & 70.89             & & 0.273             & 0.863             & 0.252             & 0.691             \\
        FLUX.1 Fill                    & \underline{12.67} & \underline{0.734} & 17.23             & 68.31             & & \underline{0.280} & \underline{0.892} & 0.277             & 0.767             \\
        \textbf{\method{} (ours) + T5 \citep{roberts2022t5x}}                & \textbf{10.49}    & \textbf{0.752}    & \textbf{19.44}    & \textbf{73.13}    & & \textbf{0.295}    & \textbf{0.904}    & \underline{0.294} & \textbf{0.793}    \\
        \bottomrule
    \end{tabular}%
    }
    
        \vspace{-1em}
\end{table}

\textbf{Without a text encoder, \method{} outperforms its baseline on all metrics.} In the fair (no-text-encoder) setting at the top of Table~\ref{tab:main_results}, \method{} beats OminiControl-w/o-prompt on all 8 metrics, with the largest gains on image quality (FID: $20.50 \to 11.13$, PSNR: $16.45 \to 19.61$). Both methods receive identical spatial conditioning and the same compute budget; the only difference is how that conditioning is encoded. This isolates the contribution of the \method{} encoder \emph{combined with our training pipeline} (Section~\ref{sec:ablations} ablates each ingredient). The \method-ImgGen evaluation set additionally includes \emph{same-image-different-text} pairs (the \emph{add}/\emph{replace}/\emph{extract} categories), and \method{} recognizes the differently rendered class labels and produces correspondingly different generations from the same source image---verifying that the model actually reads the rendered text rather than reconstructing the image condition. Note that removing the text encoder yields large efficiency savings (Section~\ref{sec:efficiency}).

\textbf{Compared to text-prompt baselines, \method{} trades a small alignment gap for major efficiency gains; the optional \method+T5 hybrid recovers the gap.} On image quality, \method{} (no T5) is on par with or slightly better than the strongest text-prompt baselines (FID: 11.13 vs.\ FLUX.1 Fill 12.67 / OminiControl 13.50; PSNR: 19.61 vs.\ FLUX.1 Fill 17.23 / ControlNet-FLUX 18.02). On alignment, the prompt-equipped methods retain a small edge (CLIP$_{\text{txt}}$ 0.270 vs.\ 0.262; region-CLIP$_{\text{txt}}$ 0.298 vs.\ 0.279)---an expected consequence of conditioning on a single masked text label rather than a full natural-language description, and of T5 embeddings sharing representational geometry with the CLIP text encoder used in these metrics. The gap is small ($\sim$3\% on global CLIP$_{\text{txt}}$) and the \method+T5 hybrid (last row of Table~\ref{tab:main_results}) closes it: it wins or ties on 7/8 metrics across all prompted baselines while preserving the same architecture. We include \method+T5 strictly as a \emph{reference point} for fair comparison against text-prompted baselines (which have access to an extra text encoder); our primary configuration remains \method{} without T5, since adding T5 forfeits the efficiency gains in Section~\ref{sec:efficiency}.

\subsection{Multi-Region Composability}
\label{sec:multiregion}

A key advantage of the proposed spatially grounded contextual generation is natural multi-region composability: multiple labeled masks are encoded jointly in a single forward pass. The main results in Table~\ref{tab:main_results} already include 1{,}500 multi-mask test samples ($N{\in}[1,5]$ masks per image) within the averaged metrics, so the quantitative gains reported there already reflect multi-region performance. Per-mask-count breakdowns are in Fig.\ref{fig:compression}(b)(c), and qualitative demonstrations of both single-step multi-region edits and sequential multi-step edits are deferred to Appendix~\ref{app:multistep} (Fig.\ref{fig:multibox}). Baselines that support only single-region control would require $K$ sequential forward passes and suffer from accumulated errors, whereas \method{} handles all regions simultaneously without architectural changes.

\subsection{Computational Efficiency}
\label{sec:efficiency}

By eliminating the text encoder entirely and encoding all conditioning information through the \method{} encoder, \method{} yields substantial reductions in token count, token size, FLOPs, and wall-clock runtime. Table~\ref{tab:efficiency} compares baseline OminiControl w/text encoder \cite{omini} vs ours --- their only architecture difference is the text encoder.

\begin{table}[t]
    \centering
    \vspace{-1em}
    \caption{Computational cost comparison v.s. baseline with text encoder  \citep{omini}, \method{} embedding eliminates all text tokens, yielding 6--50\% token reduction and 8--44\% inference time speedup. The encoder also shrinks by ${\sim}92\%$ in parameter count (T5: 4.92B $\to$ \method{} encoder: 401.6M). Gains increase at lower resolutions where the (resolution-invariant) text-encoder cost dominates. At $512{\times}512$, the noisy/image latent has size of $32{\times}32$. Runtimes measure on an H100 with 28 steps.}
    \label{tab:efficiency}
    \resizebox{\textwidth}{!}{
    \begin{tabular}{lrrr rrr rrr}
        \toprule
        & \multicolumn{3}{c}{$256{\times}256$} & \multicolumn{3}{c}{$512{\times}512$} & \multicolumn{3}{c}{$1024{\times}1024$} \\
        \cmidrule(lr){2-4} \cmidrule(lr){5-7} \cmidrule(lr){8-10}
        Metric & OminiControl & \method{} & Impr. & OminiControl & \method{} & Impr. & OminiControl & \method{} & Impr. \\
        \midrule
        Tokens/step                              & 1{,}024       & 513         & 49.9\% & 2{,}560       & 2{,}049     & 20.0\% & 8{,}704       & 8{,}193     & 5.9\%  \\
        \quad$\llcorner$ noise tokens            & 256           & 256         & --     & 1{,}024       & 1{,}024     & --     & 4{,}096       & 4{,}096     & --     \\
        \quad$\llcorner$ image $C_I$ tokens      & 256           & 256         & --     & 1{,}024       & 1{,}024     & --     & 4{,}096       & 4{,}096     & --     \\
        \quad$\llcorner$ text $C_T$ tokens       & 512           & 0           & 99.8\% & 512           & 0           & 99.8\% & 512           & 0           & 99.8\% \\
        \midrule
        Token size (count$\times$dim)            & 2{,}130{,}688 & 36{,}864    & 98.3\% & 2{,}228{,}992 & 135{,}168   & 93.9\% & 2{,}622{,}208 & 528{,}384   & 79.8\% \\
        \midrule
        Params (encoder + transformer)           & 16.82B        & 12.30B      & 26.9\% & 16.82B        & 12.30B      & 26.9\% & 16.82B        & 12.30B      & 26.9\% \\
        \quad$\llcorner$ encoder params          & 4.92B         & 401.6M      & 91.8\% & 4.92B         & 401.6M      & 91.8\% & 4.92B         & 401.6M      & 91.8\% \\
        \quad$\llcorner$ transformer params      & 11.90B        & 11.90B      & 0.0\%  & 11.90B        & 11.90B      & 0.0\%  & 11.90B        & 11.90B      & 0.0\%  \\
        \midrule
        TFLOPs ($\times 1$ enc.\ + $\times 28$ DiT)  & 792.8     & 382.1       & 51.8\% & 2{,}121.0     & 1{,}648.4   & 22.3\% & 9{,}285.1     & 8{,}567.1   & 7.7\%  \\
        \quad$\llcorner$ encoder FLOPs ($\times 1$)  & 10.3      & 0.5         & 95.4\% & 12.0          & 1.9         & 83.9\% & 19.5          & 10.3        & 47.2\% \\
        \quad$\llcorner$ transformer FLOPs ($\times 28$)  & 782.6 & 381.6       & 51.2\% & 2{,}109.0     & 1{,}646.5   & 21.9\% & 9{,}265.7     & 8{,}556.8   & 7.6\%  \\
        \midrule
        Runtime (s, $\times 1$ enc.\ + $\times 28$ DiT)   & 1.678 & 0.936       & 44.2\% & 3.620         & 2.873       & 20.6\% & 14.366        & 13.215      & 8.0\%  \\
        \bottomrule
    \end{tabular}%
    }
    
\end{table}

\begin{figure}[t]
    \centering
    \begin{subfigure}[t]{0.32\linewidth}
        \centering
        \includegraphics[width=\linewidth]{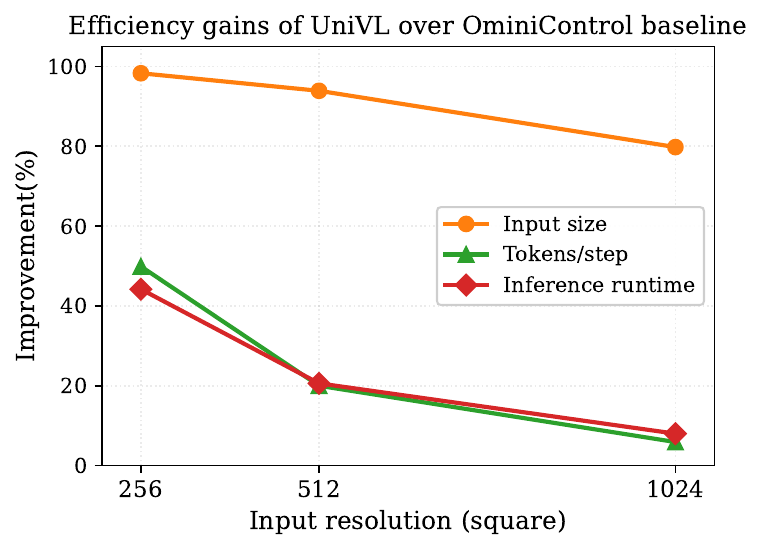}
        \caption{Efficiency gains}
        \label{fig:compression-a}
    \end{subfigure}
    \hfill
    \begin{subfigure}[t]{0.33\linewidth}
        \centering
        \includegraphics[width=\linewidth]{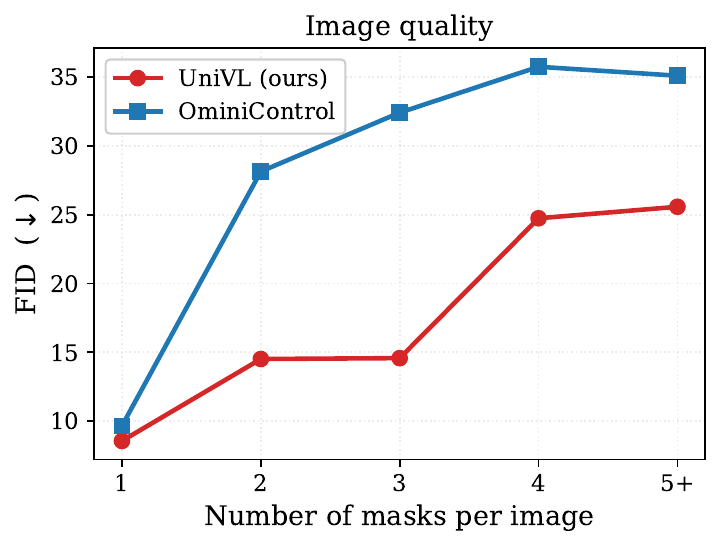}
        \caption{FID vs.\ \# masks}
        \label{fig:compression-b}
    \end{subfigure}
    \hfill
    \begin{subfigure}[t]{0.33\linewidth}
        \centering
        \includegraphics[width=\linewidth]{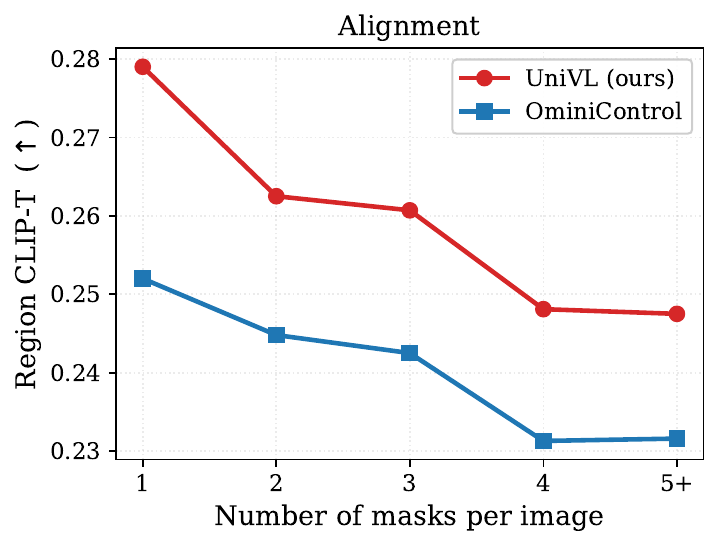}
        \caption{region-CLIP$_{\text{txt}}$ vs.\ \# masks}
        \label{fig:compression-c}
    \end{subfigure}
    \caption{(\ref{fig:compression-a})~Efficiency gains of \method{} over the baseline \cite{omini} with text encoder across image resolutions: TFLOPs and runtime savings grow at lower resolutions, from $8\%$ at $1024{\times}1024$ to $52\%$ TFLOPs / $44\%$ runtime at $256{\times}256$. (\ref{fig:compression-b})--(\ref{fig:compression-c})~Multi-region performance: \method{} (red) maintains a substantial lead over OminiControl w/o text (blue) at every mask count, both in image quality (FID, lower is better) and in region-level alignment (region-CLIP$_{\text{txt}}$, higher is better).}
    \label{fig:compression}
\end{figure}

Removing the text encoder yields large gains: token size drops 80--98\% (T5's 4096-dim embeddings dominate total element count), encoder parameters drop ${\sim}92\%$ (4.92B $\to$ 401.6M, a 26.9\% reduction in total system params), and relative TFLOPs/runtime savings scale inversely with resolution---52\%/44\% at $256^2$ down to 7.7\%/8\% at $1024^2$, since the resolution-invariant text-encoder cost dominates more when the transformer is small. Fig.\ref{fig:compression}~(a) visualizes these trends.

\subsection{Ablations and Analysis}
\label{sec:ablations}

We ablate the key design choices of the \method{} framework. All ablations are evaluated on the \method-ImgGen benchmark at $512{\times}512$ resolution unless otherwise noted.

\textbf{(a) Component ablation.}
We design four ablations (Table~\ref{tab:ablation_components}), each isolating one design choice: \emph{(R1)} replacing the DeepEncoder (SAM~+~CLIP) with the bare VAE encoder tests whether the SAM~+~CLIP architecture matters; \emph{(R2)} swapping CLIP for SigLIP (SAM retained) tests an alternative image-text-aligned backbone; \emph{(R3)} keeping CLIP but removing OCR pretraining isolates the OCR-task pretraining contribution; and \emph{(R4)} dropping $\mathcal{L}_{\text{align}}$ and training encoder+DiT jointly (with loss defined in Eq. \ref{eq:full}) tests the two-stage recipe against end-to-end training. 

\begin{table}[t!]
    \centering
    \vspace{-1em}
    \caption{Component ablation of \method{} encoder. Each row removes/replaces a single component (full configuration described in the inline summary above). The first row (our full model) is reproduced from Table~\ref{tab:main_results}.}
    \label{tab:ablation_components}
    \resizebox{\textwidth}{!}{
    \begin{tabular}{l cccc c cccc}
        \toprule
        & \multicolumn{4}{c}{Image Quality} & & \multicolumn{4}{c}{Alignment} \\
        \cmidrule(lr){2-5} \cmidrule(lr){7-10}
        Configuration & FID$\downarrow$ & SSIM$\uparrow$ & PSNR$\uparrow$ & MUSIQ$\uparrow$ & & CLIP$_{\text{txt}}\uparrow$ & CLIP$_{\text{img}}\uparrow$ & region-CLIP$_{\text{txt}}\uparrow$ & region-CLIP$_{\text{img}}\uparrow$ \\
        \midrule
        \textbf{\method{} (ours)}                            & \textbf{11.13}    & \textbf{0.754}    & \textbf{19.61}    & \textbf{73.36}    & & \textbf{0.262}    & \textbf{0.886}    & \textbf{0.279}    & \textbf{0.768}    \\
        \midrule
        (R1) \method{} encoder $\to$ VAE encoder                    & 13.60             & 0.710             & 18.34             & 72.46             & & 0.254             & 0.810             & 0.261             & 0.690             \\
        (R2) \method{} encoder  $\to$ SigLIP ViT-L                        & 13.13             & \underline{0.723} & \underline{18.80} & 72.64             & & 0.254             & 0.852             & 0.271             & 0.701             \\
                (R3) \method{} encoder  w/o OCR pretraining                         & 14.33             & 0.701             & 16.33             & 70.22             & & \underline{0.259} & 0.814             & 0.268             & 0.687             \\

        (R4) \method{} encoder  w/o Stage~1 feature align.      & \underline{13.05} & 0.698             & 18.02             & \underline{73.29} & & 0.257             & \underline{0.864} & \underline{0.274} & \underline{0.763} \\
        \bottomrule
    \end{tabular}%
    }
    
\end{table}

\emph{(R1)} \textbf{ The \method{} architecture is helpful} ($\Delta$FID $+2.5$): a pixel-reconstruction-trained VAE alone is a weaker unified VL encoder than SAM~+~CLIP. \emph{(R2)} \textbf{Substituting SigLIP yields a smaller but nontrivial gap} ($\Delta$FID $+2.0$); we did not adopt SigLIP because no public OCR-pretrained checkpoint exists for it. \emph{(R3)} \textbf{OCR-task pretraining helps the most} ($\Delta$FID $+3.2$): vanilla CLIP can detect glyphs but parses them less reliably as semantic content. R3 cannot fully separate ``OCR-task signal'' from ``additional pretraining data,'' but the smaller R2 gap (SigLIP also lacks OCR pretraining yet uses a comparably scaled image-text corpus) suggests the OCR-task aspect itself contributes a substantial portion. \emph{(R4)}\textbf{ Two-stage training outperforms end-to-end} ($\Delta$FID $+1.9$): the dedicated alignment stage produces a more stable conditioning representation before the diffusion model is exposed to it.

\textbf{(b) Other ablations/analysis.} Appendix~\ref{app:ablations} reports additional ablations: an auxiliary CLIP-loss study showing both terms are critical (Appendix~\ref{app:ablation_loss}), CLIP-loss-weight sensitivity sweeps, LoRA rank, per-resolution generation quality, and 5-seed inference stability. \method{} also outperforms the baselines on free-form masks (Appendix~\ref{app:freeform}), overlapping masks (Appendix~\ref{app:overlapping}), and zero-shot on COCO val2017 (Appendix~\ref{app:coco}); together these confirm that \method's spatial-semantic binding is a property of the conditioning representation rather than an artifact of the training mask distribution.

\section{Summary}

We introduced \emph{spatially grounded contextual image generation}, a new controllable generation task that reframes the conditioning interface from two domains (vision + language) to a single unified visual input where text is rendered onto spatial masks. Our framework, \method, replaces the text encoder with an OCR-pretrained vision encoder, achieving superior image quality over text-prompted baselines while reducing token size by up to 98\%, encoder parameters by $92\%$, TFLOPs by up to 52\%, and runtime by up to 44\%. \method{} efficiently allows specification of \emph{what} should appear \emph{where} through a single unified signal. 

\method{} has certain limitations. It is trained on rectangular masks; we report zero-shot generalization to free-form masks qualitatively in Appendix~\ref{app:freeform}, and a quantitative free-form benchmark is left to future work. \emph{Limited semantic vocabulary}: our current scope targets short phrases of a few words drawn from object-detection and image-editing annotations (1--3 words for $96\%$ of records, with a long tail up to 11 words; see Fig.\ref{fig:vocab_top40}); fully compositional natural-language sentences are not yet expressible because they would require correspondingly richer training data, and rendering long sentences inside small masks is itself constrained by mask area size. The \method{} framework is straightforward to extend along this axis through new data-augmentation pipelines (e.g., richer attribute compositions, longer instructions, or sentence-level annotations from VLM auto-labeling), which we leave to future work. \emph{Single-architecture demonstration}: we instantiate \method{} only on FLUX.1-dev; while the design (a frozen VAE-plus-CLIP encoder, a small adapter, and the two-stage training recipe) is in principle backbone-agnostic, demonstrating cross-architecture validity on other DiT-based backbones (e.g., SD3) and on U-Net-based diffusion models is left as future work. \emph{Standardized text rendering by design}: the user supplies text and our pipeline renders it through a fixed automated procedure (Courier, white-on-black, font size scaled so the rendered text spans the mask width; Appendix~\ref{app:text_render}). The encoder therefore does not need to handle arbitrary fonts or styles---inputs at training and inference are produced by the same renderer---which is a deliberate design choice rather than an unhandled distribution shift.

\clearpage
\bibliographystyle{plainnat}
\bibliography{refs}

\clearpage
\appendix
\section*{Appendix}

We provide details omitted from the main text in appendix:
\begin{itemize}[leftmargin=1.5em]
    \item \textbf{Section~\ref{app:imple}(Implementation Details):} full training hyperparameters, multi-loss formulas, mask-jittering, condition-dropout schedule, text-rendering details, and multibox training data (extends Section~\ref{sec:method}).
    \item \textbf{Section~\ref{app:inference} (Inference Details):} denoising-step  configuration and a clarification on how CLIP is used for both training and evaluation (extends Section~\ref{sec:exp_setup}).
    \item \textbf{Section~\ref{app:ocr_choice} (OCR Encoder Choice):} rationale for adapting DeepSeek-OCR's encoder, methodological comparison against alternative OCR methods, and a summary of the underlying OCR-pretraining corpus.
    \item \textbf{Section~\ref{app:ablations} (Additional Ablations):} auxiliary CLIP-loss ablation, LoRA rank, per-resolution generation quality, CLIP-loss-weight sensitivity sweeps, and 5-seed inference-stability table (extends Section~\ref{sec:ablations}).
    \item \textbf{Section~\ref{app:visualizations} (Additional Visualizations and Experiments):} zero-shot free-form masks, overlapping-mask behavior, zero-shot generalization to COCO, multi-step / single-step multibox edits, and failure cases.
    \item \textbf{Section~\ref{app:benchmark} (\method-ImgGen Benchmark Details):} per-category breakdown, augmentation pipeline, class-name vocabulary statistics, and the training/evaluation split (extends Section~\ref{sec:benchmark}).
\end{itemize}

\section{Implementation Details}
\label{app:imple}

\subsection{Training Hyperparameters}
\label{app:hparams}

Table~\ref{tab:hparams} reports the full training configuration for both stages. All training is performed in bfloat16 (Stage 1 on 2$\times$H100 80GB GPUs, Stage 2 on 6$\times$H100 80GB GPUs with gradient checkpointing); Stage 1 takes ${\approx}12.6$ hours (30K steps) and Stage 2 takes ${\approx}97.3$ hours (40K steps). The training pool is the \method-ImgGen benchmark described in Section~\ref{sec:benchmark} (${\sim}477$K unique images, ${\sim}710$K mask-text records); both stages share the same data and 512$\times$512 resolution.

\begin{table}[h]
    \centering
    \caption{Training hyperparameters for the two \method{} stages.}
    \label{tab:hparams}
   \resizebox{\textwidth}{!}{
    \begin{tabular}{lll}
        \toprule
        Hyperparameter & Stage 1 (alignment) & Stage 2 (diffusion) \\
        \midrule
        Base model                 & FLUX.1-dev (frozen DiT)                    & FLUX.1-dev (DiT~+~LoRA) \\
        Trainable params           & \method{} encoder (CLIP-LoRA + linear adapter) & DiT~LoRA + \method{} encoder (continued from Stage 1) \\
        Optimizer                  & AdamW                                       & AdamW \\
        Learning rate              & $2{\times}10^{-3}$                          & $1{\times}10^{-4}$ \\
        Weight decay               & $10^{-3}$                                   & $10^{-3}$ \\
        Batch size                 & 24                                          & 42 \\
        Max steps                  & 30K                                         & 40K \\
        Resolution                 & $512{\times}512$                            & $512{\times}512$ \\
        Precision                  & bfloat16                                    & bfloat16 \\
        Gradient checkpointing     & off                                         & on \\
        $\lambda_{\text{align}}$ (feature alignment) & 1.0                        & 1.0 \\
        $\lambda_{\text{clip-img}}$ (CLIP image)     & 0.5 (warm-up)              & 1.0 \\
        $\lambda_{\text{clip-txt}}$ (CLIP text)      & 0.0 (off)                  & 0.8 \\
        $\lambda_{\text{diff}}$ (diffusion MSE)      & N/A                        & 1.0 \\
        Multibox batch ratio       & 20\%                                        & 20\% \\
        Condition dropout (image / VAE / CLIP)  & 0.1 / 0.1 / 0.1                  & 0.1 / 0.1 / 0.1 \\
        LoRA rank ($r$, $\alpha$)  & 4, 4 (\method{} encoder)                      & 4, 4 (DiT) + 4, 4 (\method{} encoder) \\
        \bottomrule
    \end{tabular}%
    }
    
\end{table}

\subsection{Multi-Loss Supervision}
\label{app:loss}

Alongside the primary training objective of each stage---$\mathcal{L}_{\text{align}}$ in Stage~1 and the diffusion MSE $\mathcal{L}_{\text{diff}}$ in Stage~2---we apply auxiliary losses to anchor the embedding to the CLIP visual-semantic space. In Stage~1, only $\mathcal{L}_{\text{clip-img}}$ is active at a warm-up weight ($\lambda_{\text{clip-img}}{=}0.5$); in Stage~2, both auxiliary CLIP losses ($\mathcal{L}_{\text{clip-img}}{=}1.0$, $\mathcal{L}_{\text{clip-txt}}{=}0.8$) plus the feature-alignment objective remain active alongside the diffusion MSE:

\begin{enumerate}[leftmargin=1.5em]
    \item \textbf{Feature alignment} ($\mathcal{L}_{\text{align}}$, $\lambda{=}1.0$): per-token spatial fidelity between the \method{} embedding and the target VAE latent, $\|f_{\text{VL}} - z_0\|_2^2$ where $z_0 = \mathcal{E}(X)$.
    \item \textbf{CLIP image loss} ($\mathcal{L}_{\text{clip-img}}$, $\lambda{=}1.0$): enforces that \method{} features match CLIP vision patch features within the mask, $\|f_{\text{VL}} \odot m - f_{\text{CLIP}} \odot m\|^2$.
    \item \textbf{CLIP text loss} ($\mathcal{L}_{\text{clip-txt}}$, $\lambda{=}0.8$): cosine similarity between the projected pooled masked condition and the CLIP text embedding of the class name, $1 - \cos(g(\bar{c}_m),\, \text{CLIP}_{\text{text}}(\ell))$, providing category-level semantic grounding.
\end{enumerate}

The L2 loss provides instance-level alignment (reconstruct \emph{this specific object}), while the CLIP text loss provides category-level grounding (\emph{this region should be a rose}). Together they bridge the gap between \method{} visual features and semantic generation.

\subsection{Mask Jittering}
\label{app:jitter}

During training, bounding boxes are randomly perturbed:
\begin{itemize}[leftmargin=1.5em]
    \item Scale range: $(0.85, 1.5)\times$ centered on the bounding box.
    \item Translation: up to 5\% of bounding box size in each direction.
    \item Applied to both text rendering coordinates and the normalized bounding box mask.
\end{itemize}
This makes the model robust to mask size variations at inference. Without jittering, we observe a hard quality threshold at approximately 100 tokens, below which generation quality collapses.

\subsection{Condition Dropout}
\label{app:dropout}

Three independent dropout mechanisms (each with probability $p{=}0.1$) enable classifier-free guidance at inference:

\begin{table}[h]
    \centering
    \caption{Condition dropout modes during training.}
    \begin{tabular}{lll}
        \toprule
        Dropout mode & What is zeroed & Inference use \\
        \midrule
        Full image dropout      & Entire condition image       & Text-only classifier-free guidance \\
        Full condition dropout  & VAE tokens (keep CLIP)        & \texttt{image\_guidance\_scale} \\
        CLIP dropout            & CLIP tokens (keep VAE)        & \texttt{clip\_guidance\_scale} \\
        \bottomrule
    \end{tabular}
\end{table}

CLIP dropout is particularly important: it teaches the model the difference between ``with CLIP semantic instruction'' and ``without,'' enabling CLIP-specific classifier-free guidance at inference. Our inference experiments reported in the main paper do not change these guidance scales (we use the default $1.0$).

\subsection{Text Rendering on Mask}
\label{app:text_render}

Text rendering is fully automated and standardized: the user provides only the textual instruction, and our pipeline renders it with a fixed font (Courier), white-on-black, with the font size scaled so the rendered text spans the mask width (shrink until $\text{text\_width} \approx \text{bbox\_width}$). For larger masks the same line is tiled vertically to maximize text-as-pixels coverage for the \method{} encoder. Because both training and inference use this single renderer, the \method{} encoder is never asked to generalize across fonts or styles---a deliberate design choice that lets the encoder specialize on a consistent text-pixel distribution.

\subsection{Multibox Training Data}
\label{app:multibox}

20\% of training batches use multibox data: images with 2--5 non-overlapping masked regions, each with a different class label. Mask sampling rules:
\begin{itemize}[leftmargin=1.5em]
    \item Each mask area: 4\%--40\% of image area.
    \item Pairwise IoU $< 0.05$ between all masks.
    \item If the non-overlapping constraint fails after 5 retries: fall back to 2 masks, then 1.
\end{itemize}

\section{Inference Details}
\label{app:inference}

At inference, we use 28 denoising steps with \texttt{guidance\_scale=1.0} (no text classifier-free guidance, since no text prompt is used). The default configuration uses \texttt{image\_guidance\_scale=1.0} and \texttt{clip\_guidance\_scale=1.0} (both off), requiring only a single forward pass per step. We did not vary these guidance scales in the paper's experiments since they yielded no significant performance improvement.

\paragraph{Metrics clarification.}
We use CLIP in three places: dataset filtering (Section~\ref{sec:benchmark}), auxiliary training losses ($\mathcal{L}_{\text{clip-img}}$, $\mathcal{L}_{\text{clip-txt}}$), and evaluation metrics (CLIP$_{\text{txt}}$, CLIP$_{\text{img}}$, region-CLIP$_{\text{txt}}$, region-CLIP$_{\text{img}}$). To avoid any concern of an unfair advantage, we note that the CLIP encoder is treated as a \emph{frozen function} throughout training and evaluation: its parameters are never updated, and it is applied identically to every method we evaluate. As a result, there is no risk of CLIP overfitting to our model in the conventional sense---it is the same fixed metric used to evaluate the baselines. Additionally, half of our reported metrics (FID, SSIM, PSNR, MUSIQ) are CLIP-free and our method also leads on these, suggesting the gains are not specific to CLIP-based scoring.

\section{OCR Encoder Choice}
\label{app:ocr_choice}

We provide additional rationale for using the DeepSeek-OCR encoder~\citep{deepseekocr} as the backbone of \method, and discuss why we did not adopt other OCR-capable vision models.

\paragraph{Why DeepSeek-OCR.} The encoder of DeepSeek-OCR (called \emph{DeepEncoder}) combines a SAM-base window-attention block with a CLIP ViT-L global-attention block, separated by a 16$\times$ convolutional compressor. This design has three properties we rely on: (i)~the CLIP block produces image-text-aligned semantic features, which are important for reading rendered glyphs and binding them to spatial locations; (ii)~the SAM block adds dense local perception, which helps with the small text regions inside masks; and (iii)~the encoder has been pretrained on a large and diverse OCR corpus (${\sim}30$M PDF pages, scene OCR, and general vision data), giving the encoder broad text-as-pixels recognition out of the box. Other components of the original DeepSeek-OCR system (multiple resolution modes, the DeepSeek-3B-MoE decoder) are not needed for our task; we use only the encoder, in a single native-resolution mode, since our masks are natural-image-scale rather than multi-page document scans. A successor system (DeepSeek-OCR2,~\citet{deepseekocr2}) explores larger encoders and richer multi-task pretraining; the conditioning idea in \method{} is encoder-agnostic and we expect it to apply with stronger encoders, but for this paper we adopt the minimalist DeepSeek-OCR encoder for simplicity and reproducibility.

\paragraph{Summary of OCR encoder pretraining.} For completeness, we summarize the data used to pretrain the DeepSeek-OCR encoder before our feature-reconstruction stage; the original training was performed by~\citet{deepseekocr} and we use their released checkpoint as-is. The pretraining mix is roughly \emph{70\% OCR / 20\% general vision / 10\% text-only}. The OCR portion comprises two regimes:
\begin{itemize}[leftmargin=1.5em]
    \item \textbf{Document OCR} (${\sim}33$M pages): ${\sim}30$M PDF pages spanning ${\sim}100$ languages (${\sim}25$M Chinese, 5M English, plus minority languages), labeled with both \emph{coarse} text extracted via \texttt{fitz} and \emph{fine} layout-and-text annotations (2M Chinese + 2M English from PP-DocLayout/MinerU/GOT-OCR2.0; ${\sim}600$K minority-language patches). An additional ${\sim}3$M Word-document pairs cover formulas and HTML tables without layout.
    \item \textbf{Natural-scene OCR} (${\sim}20$M images): 10M Chinese + 10M English from LAION and Wukong, labeled with PaddleOCR.
\end{itemize}
The remaining 20\% general-vision data preserves a generic vision interface (caption/detection/grounding) and the 10\% text-only data preserves language modeling capability. This pretraining gives the encoder broad text-as-pixels recognition that we then adapt for our generation conditioning task in Stage~1 of our pipeline (Section~\ref{sec:method}).

\paragraph{Why not other OCR methods.} Two alternative public OCR-capable vision models exist:
\begin{itemize}[leftmargin=1.5em]
    \item \textbf{Nougat}~\citep{blecher2023nougat}: end-to-end OCR method. Pretrained on a much narrower domain (arXiv-style PDFs only) and at smaller scale than DeepSeek-OCR. Adopting Nougat would limit out-of-distribution robustness to the natural-image-scale, freely-rendered text we use in the contextual condition.
    \item \textbf{GOT-OCR2.0}~\citep{wei2024gotocr}: another end-to-end OCR method covering charts, formulas, and document parsing. Comparable scope to DeepSeek-OCR but with a smaller pretraining corpus and a tighter coupling to its decoder. The encoder is less easily decoupled for use as a standalone conditioning encoder, and there is no published encoder-only checkpoint.
\end{itemize}
We chose DeepSeek-OCR primarily for the scale and diversity of its pretraining data, and because its encoder is publicly released as a standalone module, making it convenient to bolt onto FLUX without re-running large-scale OCR pretraining.
\section{Additional Ablations}
\label{app:ablations}

We report additional ablations that complement Section~\ref{sec:ablations} in the main paper. All experiments use the default configuration (resolution $512{\times}512$, LoRA $r{=}4$, $\lambda_{\text{clip-img}}{=}1.0$, $\lambda_{\text{clip-txt}}{=}0.8$, 40K Stage-2 training steps) unless otherwise noted, and the reference row (``ours'') uses our final model as reported in Table~\ref{tab:main_results}.

\subsection{Auxiliary CLIP Loss}
\label{app:ablation_loss}

We study the effect of adding CLIP-based auxiliary losses on top of the feature alignment objective in Stage~1. We compare four configurations: feature alignment only ($\mathcal{L}_{\text{align}}$), adding a CLIP image-space loss ($\mathcal{L}_{\text{align}} + \mathcal{L}_{\text{clip-img}}$), adding only a CLIP text loss ($\mathcal{L}_{\text{align}} + \mathcal{L}_{\text{clip-txt}}$), and adding both ($\mathcal{L}_{\text{align}} + \mathcal{L}_{\text{clip-img}} + \mathcal{L}_{\text{clip-txt}}$, ours). Results on the \method-ImgGen benchmark are shown in Table~\ref{tab:ablation_loss}.

\begin{table}[h]
    \centering
    \caption{Auxiliary CLIP loss ablation on top of the Stage~1 feature alignment objective. The full configuration ($\mathcal{L}_{\text{align}} + \mathcal{L}_{\text{clip-img}} + \mathcal{L}_{\text{clip-txt}}$, ours) leads on all 8 metrics; the CLIP image and text losses act as complementary regularizers for image quality and region-level alignment, respectively. Best in \textbf{bold}, second-best \underline{underlined}.}
    \label{tab:ablation_loss}
    \resizebox{\textwidth}{!}{
    \begin{tabular}{l cccc c cccc}
        \toprule
        & \multicolumn{4}{c}{Image Quality} & & \multicolumn{4}{c}{Alignment} \\
        \cmidrule(lr){2-5} \cmidrule(lr){7-10}
        Configuration & FID$\downarrow$ & SSIM$\uparrow$ & PSNR$\uparrow$ & MUSIQ$\uparrow$ & & CLIP$_{\text{txt}}\uparrow$ & CLIP$_{\text{img}}\uparrow$ & region-CLIP$_{\text{txt}}\uparrow$ & region-CLIP$_{\text{img}}\uparrow$ \\
        \midrule
        No CLIP loss ($\mathcal{L}_{\text{align}}$ only)                  & 28.11             & 0.665             & 17.97             & 69.36             & & \underline{0.256} & 0.791             & \underline{0.277} & 0.709             \\
        $+ \mathcal{L}_{\text{clip-img}}$                                 & 21.77             & 0.663             & 17.12             & \underline{73.03} & & 0.251             & 0.840             & 0.273             & 0.746             \\
        $+ \mathcal{L}_{\text{clip-txt}}$                                 & \underline{12.72} & \underline{0.724} & \underline{18.80} & 72.61             & & 0.253             & \underline{0.876} & \underline{0.277} & \underline{0.751} \\
        \textbf{$+ \mathcal{L}_{\text{clip-img}} + \mathcal{L}_{\text{clip-txt}}$ (ours)} & \textbf{11.13} & \textbf{0.754} & \textbf{19.61} & \textbf{73.36} & & \textbf{0.262} & \textbf{0.886} & \textbf{0.279} & \textbf{0.768} \\
        \bottomrule
    \end{tabular}%
    }
    
\end{table}

The two CLIP losses provide complementary regularization. The CLIP image loss adds a perceptual prior ($\Delta$FID $-6.3$, $\Delta$MUSIQ $+3.7$ over no-CLIP), preventing the visually flat outputs of pure feature alignment. The CLIP text loss adds a category-level anchor ($\Delta$FID $-15.4$ over no-CLIP, $\Delta$region-CLIP$_{\text{img}}$ $+0.04$), preventing the adapter from drifting to off-category representations. Combining both yields the strongest configuration; notably, $+\mathcal{L}_{\text{clip-txt}}$ alone (FID $12.72$) already outperforms $+\mathcal{L}_{\text{clip-img}}$ alone (FID $21.77$).

\subsection{CLIP Image Loss Weight}
\label{app:ablation_lambda_img}

We complement Appendix~\ref{app:ablation_loss} (which studies the \emph{presence} of the auxiliary CLIP losses) with a sensitivity analysis on the weight of the CLIP image loss, $\lambda_{\text{clip-img}}$, holding the CLIP text loss weight fixed at $\lambda_{\text{clip-txt}}{=}0.8$.

\begin{table}[h]
    \centering
    \caption{Sensitivity to the CLIP image loss weight $\lambda_{\text{clip-img}}$.}
    \resizebox{\textwidth}{!}{
    \begin{tabular}{l cccc c cccc}
        \toprule
        & \multicolumn{4}{c}{Image Quality} & & \multicolumn{4}{c}{Alignment} \\
        \cmidrule(lr){2-5} \cmidrule(lr){7-10}
        $\lambda_{\text{clip-img}}$ & FID$\downarrow$ & SSIM$\uparrow$ & PSNR$\uparrow$ & MUSIQ$\uparrow$ & & CLIP$_{\text{txt}}\uparrow$ & CLIP$_{\text{img}}\uparrow$ & region-CLIP$_{\text{txt}}\uparrow$ & region-CLIP$_{\text{img}}\uparrow$ \\
        \midrule
        0.1 & 13.81 & 0.732 & 19.42 & 72.64 && 0.238 & 0.851 & 0.255 & 0.736 \\
        \textbf{1.0} & \textbf{11.13} & \textbf{0.754} & \textbf{19.61} & \textbf{73.36} & & \textbf{0.262} & \textbf{0.886} & \textbf{0.279} & \textbf{0.768} \\
        2.0                 & 15.72          & 0.665          & 17.53          & 73.36          & & 0.255          & 0.854          & 0.274          & 0.756          \\
        \bottomrule
    \end{tabular}%
    }
    
\end{table}

Increasing $\lambda_{\text{clip-img}}$ pulls the \method{} adapter too strongly toward pooled CLIP-space similarity at the cost of pixel-level fidelity, while too small a value (0.1) under-regularizes. $\lambda_{\text{clip-img}}{=}1.0$ balances feature alignment with semantic regularization from CLIP image space.

\subsection{CLIP Text Loss Weight}
\label{app:ablation_lambda_txt}

We additionally sweep the CLIP text loss weight $\lambda_{\text{clip-txt}}$ across three values, holding $\lambda_{\text{clip-img}}{=}1.0$ fixed.

\begin{table}[h]
    \centering
    \caption{Sensitivity to the CLIP text loss weight $\lambda_{\text{clip-txt}}$. }
    \resizebox{\textwidth}{!}{
    \begin{tabular}{l cccc c cccc}
        \toprule
        & \multicolumn{4}{c}{Image Quality} & & \multicolumn{4}{c}{Alignment} \\
        \cmidrule(lr){2-5} \cmidrule(lr){7-10}
        $\lambda_{\text{clip-txt}}$ & FID$\downarrow$ & SSIM$\uparrow$ & PSNR$\uparrow$ & MUSIQ$\uparrow$ & & CLIP$_{\text{txt}}\uparrow$ & CLIP$_{\text{img}}\uparrow$ & region-CLIP$_{\text{txt}}\uparrow$ & region-CLIP$_{\text{img}}\uparrow$ \\
        \midrule
        0.1                 & 14.43             & 0.737             &    19.78          &71.75            & & 0.246             & 0.868             &   0.261        &0.748             \\

        \textbf{0.8 (ours)} & \textbf{11.13} & \textbf{0.754} & \textbf{19.61} & \textbf{73.36} & & \textbf{0.262} & \textbf{0.886} & \textbf{0.279} & \textbf{0.768} \\
        2.0& 14.28            & 0.739             &    19.74          &71.89            & & 0.242             & 0.859             &   0.262        &0.745  \\
        \bottomrule
    \end{tabular}%
    }
    
\end{table}

The CLIP text loss anchors the pooled masked condition to the CLIP text embedding of the class label, providing category-level semantic grounding. Too small a weight leaves the adapter free to drift away from the labeled category; too large a weight may over-constrain the embedding and reduce sample diversity.

\subsection{LoRA Rank}

We compare LoRA ranks $r{=}4$ and $r{=}8$ for the DiT fine-tuning stage (Stage~2).

\begin{table}[h]
    \centering
    \caption{LoRA rank ablation. Doubling rank from 4 to 8 doubles trainable parameters without improvement across any metric. Best in \textbf{bold}.}
    \resizebox{\textwidth}{!}{
    \begin{tabular}{l c cccc c cccc}
        \toprule
        & Params & \multicolumn{4}{c}{Image Quality} & & \multicolumn{4}{c}{Alignment} \\
        \cmidrule(lr){3-6} \cmidrule(lr){8-11}
        LoRA rank & (rel.) & FID$\downarrow$ & SSIM$\uparrow$ & PSNR$\uparrow$ & MUSIQ$\uparrow$ & & CLIP$_{\text{txt}}\uparrow$ & CLIP$_{\text{img}}\uparrow$ & region-CLIP$_{\text{txt}}\uparrow$ & region-CLIP$_{\text{img}}\uparrow$ \\
        \midrule
        \textbf{$r{=}4$ (ours)} & $1\times$ & \textbf{11.13} & \textbf{0.754} & \textbf{19.61} & \textbf{73.36} & & \textbf{0.262} & \textbf{0.886} & \textbf{0.279} & \textbf{0.768} \\
        $r{=}8$                 & $2\times$ & 15.77          & 0.669          & 17.62          & 73.20          & & 0.255          & 0.855          & 0.275          & 0.754          \\
        \bottomrule
    \end{tabular}%
    }
    
\end{table}

$r{=}4$ matches or exceeds $r{=}8$ on all metrics, suggesting the bottleneck is not DiT LoRA capacity but the \method{} adapter, which already produces a sufficiently expressive \method{} embedding in Stage~1.

\subsection{Generation Quality Across Resolutions}
\label{app:ablation_resolution}

Table~\ref{tab:efficiency} (main paper) quantifies the efficiency gains of \method{} at $256{\times}256$, $512{\times}512$, and $1024{\times}1024$. To confirm that these gains do not come at the cost of generation quality, we train separate \method{} models at each resolution and report their generation metrics in Table~\ref{tab:ablation_resolution}. We emphasize that these rows are \emph{not} directly comparable: FID, PSNR, SSIM, and CLIP-based metrics all depend on the input resolution (e.g., FID is sensitive to the Inception network's effective receptive field, and SSIM/PSNR depend on the pixel grid). The table therefore serves as a per-resolution sanity check showing that \method{} produces reasonable outputs at each scale, \emph{not} as a cross-resolution comparison.

\begin{table}[h]
    \centering
    \caption{Per-resolution generation quality for \method. Each row is evaluated at its own resolution; metrics are not directly comparable across rows (FID, PSNR, and SSIM all depend on input resolution). The purpose of this table is to validate that \method{} maintains reasonable generation quality at every scale reported in the efficiency analysis (Table~\ref{tab:efficiency}).}
    \label{tab:ablation_resolution}
    \resizebox{\textwidth}{!}{
    \begin{tabular}{l cccc c cccc}
        \toprule
        & \multicolumn{4}{c}{Image Quality} & & \multicolumn{4}{c}{Alignment} \\
        \cmidrule(lr){2-5} \cmidrule(lr){7-10}
        Resolution & FID$\downarrow$ & SSIM$\uparrow$ & PSNR$\uparrow$ & MUSIQ$\uparrow$ & & CLIP$_{\text{txt}}\uparrow$ & CLIP$_{\text{img}}\uparrow$ & region-CLIP$_{\text{txt}}\uparrow$ & region-CLIP$_{\text{img}}\uparrow$ \\
        \midrule
        $256{\times}256$                 & 17.55 & 0.632 & 17.15 & 54.66 & & 0.254 & 0.830 & 0.277 & 0.761 \\
        $512{\times}512$ (default)       & 11.13 & 0.754 & 19.61 & 73.36 & & 0.262 & 0.886 & 0.279 & 0.768 \\
        $1024{\times}1024$               & 10.87 & 0.749 & 20.57 & 73.83 & & 0.254 & 0.855 & 0.271 & 0.757 \\
        \bottomrule
    \end{tabular}%
    }
    
\end{table}

At all three resolutions \method{} produces outputs with coherent image quality and alignment---confirming that the efficiency gains reported in Table~\ref{tab:efficiency} (up to 98\% token-size reduction and 52\% TFLOPs reduction at $256{\times}256$) are not bought at the cost of broken generations. The $256{\times}256$ model's low MUSIQ score is consistent with the resolution itself: fine textures are simply not resolvable at this scale, and this limit applies to any generator, not \method{} specifically.

\subsection{Robustness to Inference Seed}
\label{app:std}

To assess statistical significance of the headline comparison in Table~\ref{tab:main_results}, we re-run inference 5 times with different seeds (using the same trained checkpoint and the same 3{,}000-sample evaluation set) and report the per-cell standard deviations alongside the means in Table~\ref{tab:main_results_std}. The dispersion is small (e.g., $\sigma_{\text{FID}}\le 0.53$ across all rows), and the rankings reported in the main paper are preserved at every metric.

\begin{table}[h]
    \centering
    \caption{Mean $\pm$ standard deviation across 5 inference seeds on the \method-ImgGen benchmark. Means are identical to Table~\ref{tab:main_results} (reproduced for convenience).}
    \label{tab:main_results_std}
    \resizebox{\textwidth}{!}{
    \begin{tabular}{l cccc c cccc}
        \toprule
        & \multicolumn{4}{c}{Image Quality} & & \multicolumn{4}{c}{Alignment} \\
        \cmidrule(lr){2-5} \cmidrule(lr){7-10}
        Method & FID$\downarrow$ & SSIM$\uparrow$ & PSNR$\uparrow$ & MUSIQ$\uparrow$ & & CLIP$_{\text{txt}}\uparrow$ & CLIP$_{\text{img}}\uparrow$ & region-CLIP$_{\text{txt}}\uparrow$ & region-CLIP$_{\text{img}}\uparrow$ \\
        \midrule
        \multicolumn{10}{l}{\emph{Without text encoder}} \\
        ControlNet-FLUX                  & 13.98$\pm$0.40 & 0.734$\pm$0.003 & 18.22$\pm$0.14 & 70.50$\pm$0.08 & & 0.250$\pm$0.005 & 0.856$\pm$0.005 & 0.249$\pm$0.009 & 0.689$\pm$0.011 \\
        OminiControl~\citep{omini}       & 20.50$\pm$0.36 & 0.719$\pm$0.003 & 16.45$\pm$0.19 & 72.05$\pm$0.15 & & 0.249$\pm$0.005 & 0.849$\pm$0.013 & 0.271$\pm$0.006 & 0.742$\pm$0.021 \\
        \method{} (ours)                   & 11.13$\pm$0.33 & 0.754$\pm$0.002 & 19.61$\pm$0.15 & 73.36$\pm$0.09 & & 0.262$\pm$0.001 & 0.886$\pm$0.010 & 0.279$\pm$0.006 & 0.768$\pm$0.017 \\
        \midrule
        \multicolumn{10}{l}{\emph{With text encoder}} \\
        ControlNet~\citep{zhang2023controlnet} & 23.93$\pm$0.53 & 0.684$\pm$0.003 & 15.30$\pm$0.25 & 69.82$\pm$0.13 & & 0.264$\pm$0.007 & 0.859$\pm$0.011 & 0.282$\pm$0.010 & 0.763$\pm$0.021 \\
        OminiControl~\citep{omini}             & 13.50$\pm$0.42 & 0.718$\pm$0.003 & 16.13$\pm$0.20 & 72.81$\pm$0.11 & & 0.270$\pm$0.005 & 0.871$\pm$0.014 & 0.298$\pm$0.009 & 0.770$\pm$0.031 \\
        ControlNet-FLUX                        & 13.77$\pm$0.31 & 0.731$\pm$0.002 & 18.02$\pm$0.19 & 70.89$\pm$0.07 & & 0.273$\pm$0.004 & 0.863$\pm$0.009 & 0.252$\pm$0.011 & 0.691$\pm$0.021 \\
        FLUX.1 Fill (fine-tuned)               & 12.67$\pm$0.36 & 0.734$\pm$0.002 & 17.23$\pm$0.11 & 68.31$\pm$0.08 & & 0.280$\pm$0.004 & 0.892$\pm$0.003 & 0.277$\pm$0.007 & 0.767$\pm$0.010 \\
        \method{} (ours) + T5                    & 10.49$\pm$0.41 & 0.752$\pm$0.003 & 19.44$\pm$0.16 & 73.13$\pm$0.11 & & 0.295$\pm$0.002 & 0.904$\pm$0.009 & 0.294$\pm$0.008 & 0.793$\pm$0.016 \\
        \bottomrule
    \end{tabular}%
    }
    
\end{table}

\section{Additional Visualizations and Experiments}
\label{app:visualizations}

\subsection{Non-Square (Free-Form) Masks}
\label{app:freeform}

\method{} is trained exclusively on axis-aligned rectangular masks, but the underlying task formulation places no architectural constraint on mask shape---the contextual condition $C_I$ is just an image with a region masked and a text label rendered inside it. We test \method's behavior on irregular, free-form masks at inference time \emph{without any retraining or fine-tuning} (zero-shot evaluation).

\begin{figure}[t!]
    \centering
    \includegraphics[width=\linewidth]{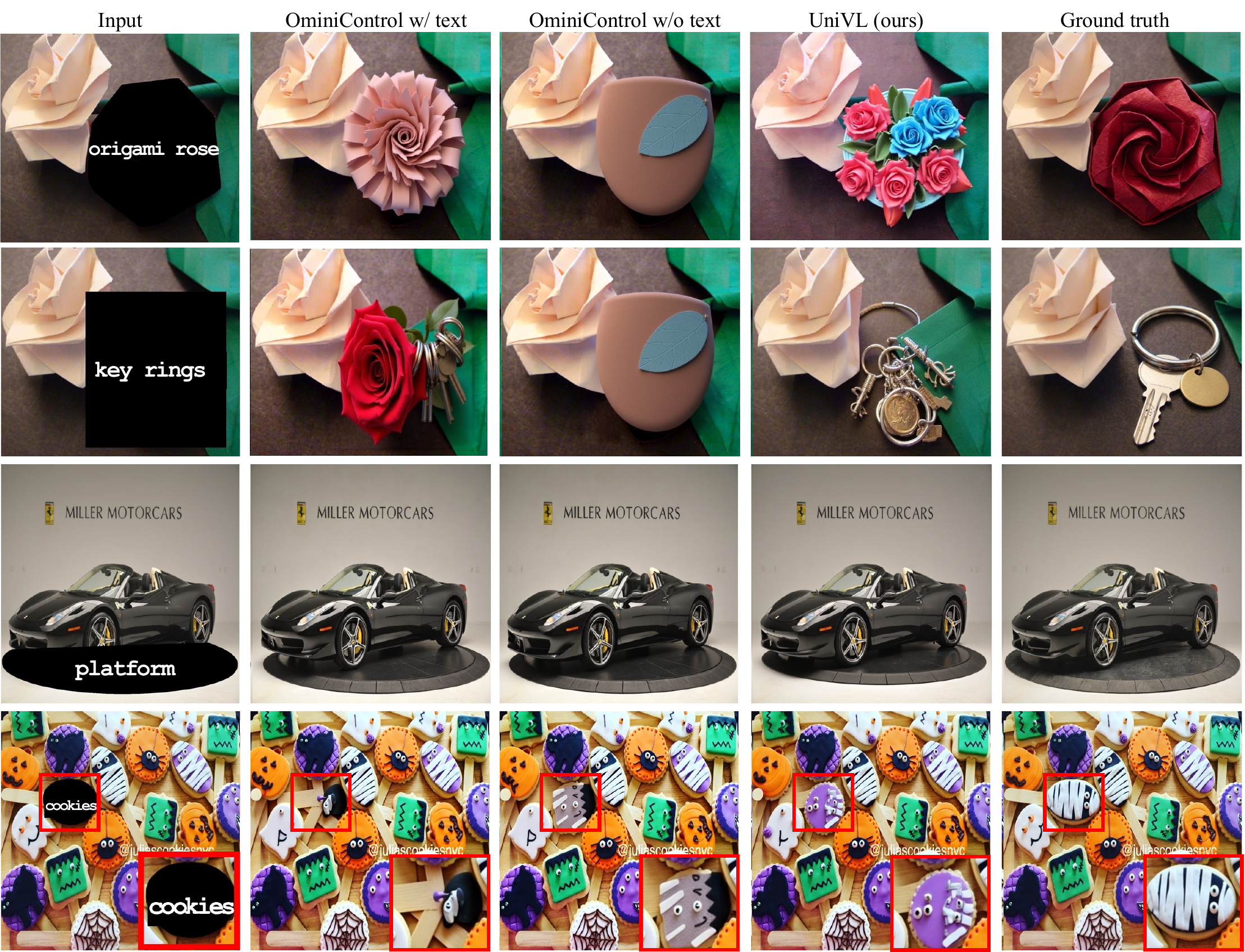}
    \caption{Zero-shot test on irregular mask shape. \emph{Row 1:} \method{} generates an image whose foreground content fills the irregular mask shape, matching the text label rendered inside the non-rectangular region. \emph{Row 4:} \method{} respects the round-shape cookie mask and outperforms the OminiControl baseline, which defaults to a bounding-rectangle fill and ignores the irregular boundary. Although \method{} is trained only on rectangular masks, at inference time it respects both the user-provided mask shape and the semantic instruction, indicating that \method's spatial-semantic binding is a property of the conditioning representation rather than an artifact of the training mask distribution.}
    \label{fig:freeform}
\end{figure}

Two findings emerge: (i)~the \method{} encoder reliably reads text rendered inside non-rectangular regions, and (ii)~the diffusion model respects the actual mask boundary rather than the bounding rectangle, generating content that conforms to the user's intended shape. This zero-shot generalization confirms that \method's design does not assume rectangular masks; extending training to free-form masks is expected to further improve fidelity at irregular boundaries.

\paragraph{Quantitative evaluation on free-form masks.} To complement the qualitative results above, we evaluate all three methods on a held-out free-form mask test set (irregular polygonal masks; same image distribution and prompts as the in-domain set). Table~\ref{tab:freeform} reports the metrics.

\begin{table}[h]
    \centering
    \caption{Zero-shot quantitative results on free-form (non-rectangular) masks. \method{} is best on 5/8 metrics (FID, SSIM, PSNR, CLIP$_{\text{img}}$, region-CLIP$_{\text{img}}$) and second-best on the two text-alignment metrics. Best in \textbf{bold}, second-best \underline{underlined}.}
    \label{tab:freeform}
    \resizebox{\textwidth}{!}{
    \begin{tabular}{l cccc c cccc}
        \toprule
        & \multicolumn{4}{c}{Image Quality} & & \multicolumn{4}{c}{Alignment} \\
        \cmidrule(lr){2-5} \cmidrule(lr){7-10}
        Method & FID$\downarrow$ & SSIM$\uparrow$ & PSNR$\uparrow$ & MUSIQ$\uparrow$ & & CLIP$_{\text{txt}}\uparrow$ & CLIP$_{\text{img}}\uparrow$ & region-CLIP$_{\text{txt}}\uparrow$ & region-CLIP$_{\text{img}}\uparrow$ \\
        \midrule
        \textbf{\method{} (ours)} & \textbf{8.64}     & \textbf{0.8376}    & \textbf{23.34}     & 72.82             & & \underline{0.2619} & \textbf{0.9486}    & \underline{0.2872} & \textbf{0.8177} \\
        OminiControl w/o text   & 10.44             & 0.7833             & \underline{21.52}  & \textbf{73.20}    & & 0.2599             & 0.9298             & 0.2867             & 0.7883          \\
        OminiControl w/ text    & \underline{9.97}  & \underline{0.7838} & 21.21              & \underline{73.08} & & \textbf{0.2686}    & \underline{0.9375} & \textbf{0.2904}    & \underline{0.8018} \\
        \bottomrule
    \end{tabular}%
    }
    
\end{table}

Two observations follow. First, \emph{\method{} continues to outperform the OminiControl baseline} on free-form masks, even though the model has never seen non-rectangular masks during training: the headline rankings from the in-domain rectangular benchmark (Table~\ref{tab:main_results}) are preserved zero-shot. Second, \emph{free-form-mask metrics are substantially better than rectangular-mask metrics for \method{} on image quality}---FID drops from $11.13 \to 8.64$, SSIM rises from $0.754 \to 0.838$, PSNR from $19.61 \to 23.34$, and CLIP$_{\text{img}}$ from $0.886 \to 0.949$. We hypothesize that the irregular mask shape itself acts as additional conditioning signal (a tighter object outline) compared to a rectangular bounding box, which provides a stronger spatial prior to the diffusion model and reduces the size of the region the model must hallucinate. This is consistent with the finding that the gap is largest on \emph{image-quality} metrics (which reward background preservation and shape fidelity) and roughly neutral on the text-alignment metrics (which reward only that the right semantic class fills the region). Extending training to free-form masks is therefore likely to widen rather than narrow this advantage.

\subsection{Overlapping Masks}
\label{app:overlapping}

\begin{figure}[h]
    \centering
    \includegraphics[width=\linewidth]{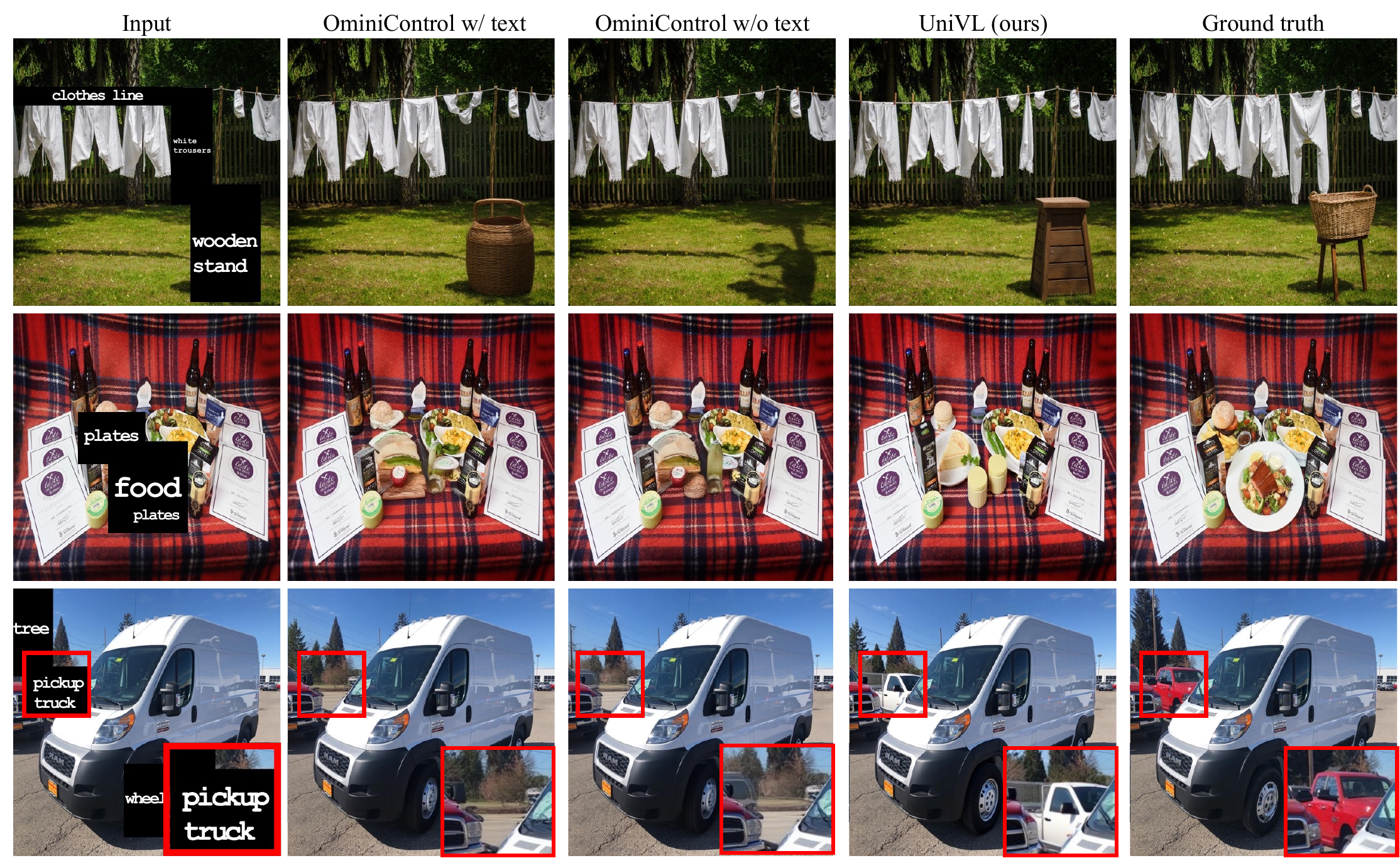}
    \caption{Overlapping masks. \emph{Row 1:} \method{} works when two masks do not overlap too much---each labeled region is generated as instructed. \emph{Row 2 (failure case):} when the ``food'' and ``plate'' boxes totally overlay, \method{} cannot disambiguate which label belongs to which region and produces a single fused object. \emph{Row 3:} compared to existing methods, \method{} follows the ``pickup truck'' instruction at the desired location, while the baselines either drop the instruction or place it outside the requested region.}
    \label{fig:overlapping_mask}
\end{figure}

We probe \method's behavior when multiple labeled masks overlap. Fig.\ref{fig:overlapping_mask} shows three representative cases.
The takeaway: \method{} handles partially overlapping masks robustly but degrades gracefully when two masks occupy nearly the same pixels---the contextual condition becomes ambiguous because two text labels are rendered into the same region, and disambiguation would require additional cues (e.g., explicit foreground/background ordering) that the current interface does not provide.

\subsection{Results on COCO}
\label{app:coco}

\begin{figure}[h]
    \centering
    \includegraphics[width=\linewidth]{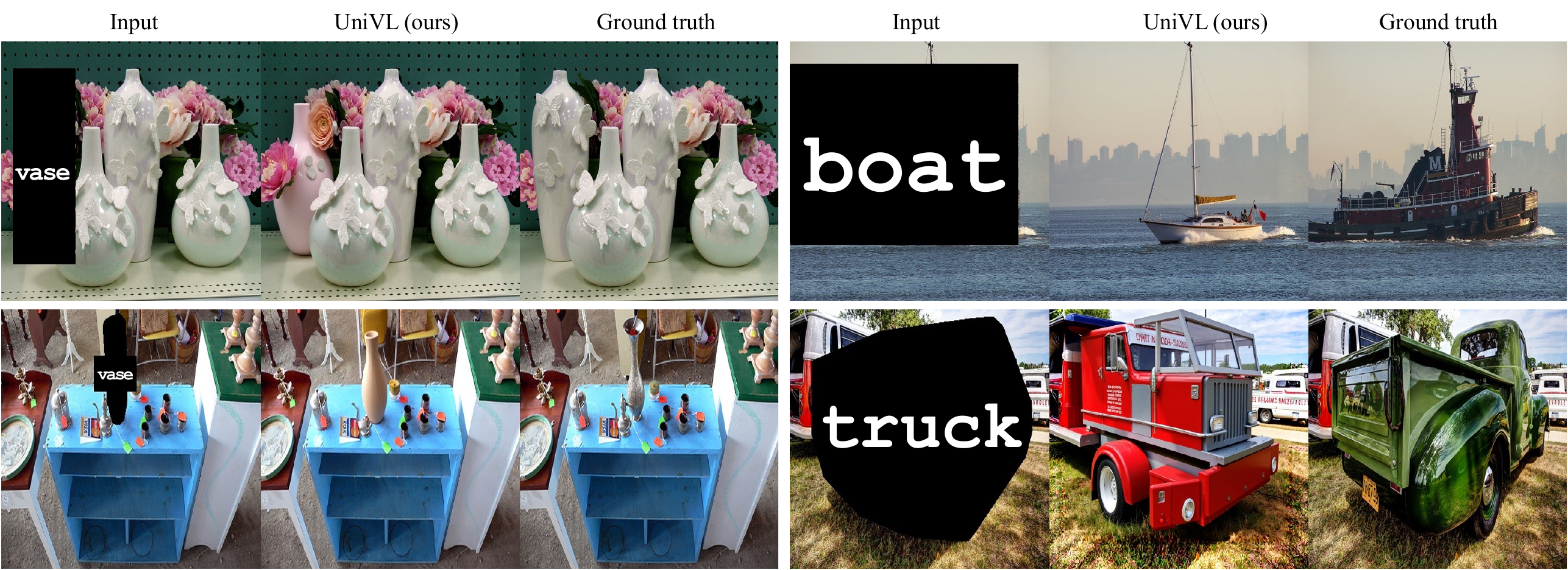}
    \caption{Zero-shot \method{} on COCO val2017. \method{} generalizes well to a held-out natural-image distribution despite never having seen COCO during training, producing region-faithful generations across diverse object categories.}
    \label{fig:coco}
\end{figure}

To assess generalization beyond the \method-ImgGen training distribution, we evaluate \method{} zero-shot on COCO val2017. We randomly sample 3{,}000 images, use the GT bounding boxes (axis-aligned squares) and class names as the spatial-text condition, and run \method{} with no fine-tuning. Table~\ref{tab:coco} compares \method{} against OminiControl with and without text prompt under identical conditioning, and Fig.\ref{fig:coco} shows representative qualitative results.

\begin{table}[h]
    \centering
    \caption{Zero-shot results on COCO val2017 (3{,}000 randomly sampled images, square masks). \method{} outperforms both OminiControl variants on FID, SSIM, PSNR, and CLIP$_{\text{img}}$ despite never having seen COCO during training. Best in \textbf{bold}, second-best \underline{underlined}.}
    \label{tab:coco}
    \resizebox{\textwidth}{!}{
    \begin{tabular}{l cccc c cccc}
        \toprule
        & \multicolumn{4}{c}{Image Quality} & & \multicolumn{4}{c}{Alignment} \\
        \cmidrule(lr){2-5} \cmidrule(lr){7-10}
        Method & FID$\downarrow$ & SSIM$\uparrow$ & PSNR$\uparrow$ & MUSIQ$\uparrow$ & & CLIP$_{\text{txt}}\uparrow$ & CLIP$_{\text{img}}\uparrow$ & region-CLIP$_{\text{txt}}\uparrow$ & region-CLIP$_{\text{img}}\uparrow$ \\
        \midrule
        \textbf{\method{} (ours)}      & \textbf{16.99}    & \textbf{0.7592}   & \textbf{20.69}    & 71.03             & & \underline{0.2638} & \textbf{0.8791}   & \underline{0.2707} & \underline{0.7305} \\
        OminiControl w/o text        & 21.37             & 0.7109            & 19.50             & \underline{71.06} & & 0.2609             & 0.8575            & 0.2635             & 0.7129             \\
        OminiControl w/ text         & \underline{17.83} & \underline{0.7135} & \underline{19.11} & \textbf{71.73}   & & \textbf{0.2810}    & \underline{0.8679} & \textbf{0.2814}   & \textbf{0.7320}    \\
        \bottomrule
    \end{tabular}%
    }
    
\end{table}

\method{} leads on FID, SSIM, PSNR, and CLIP$_{\text{img}}$ over both prompted and unprompted OminiControl. As expected, OminiControl with text prompt retains an alignment edge on text-conditioned metrics (CLIP$_{\text{txt}}$, region-CLIP$_{\text{txt}}$, region-CLIP$_{\text{img}}$) since it has access to the natural-language label as a separate signal; \method{} narrows this gap while operating in the no-text-encoder setting, consistent with the trends on the in-domain \method-ImgGen benchmark (Table~\ref{tab:main_results}).

\subsection{Multi-Step and Single-Step Multibox Edits}
\label{app:multistep}

\begin{figure}[h]
    \centering
    \includegraphics[width=\linewidth]{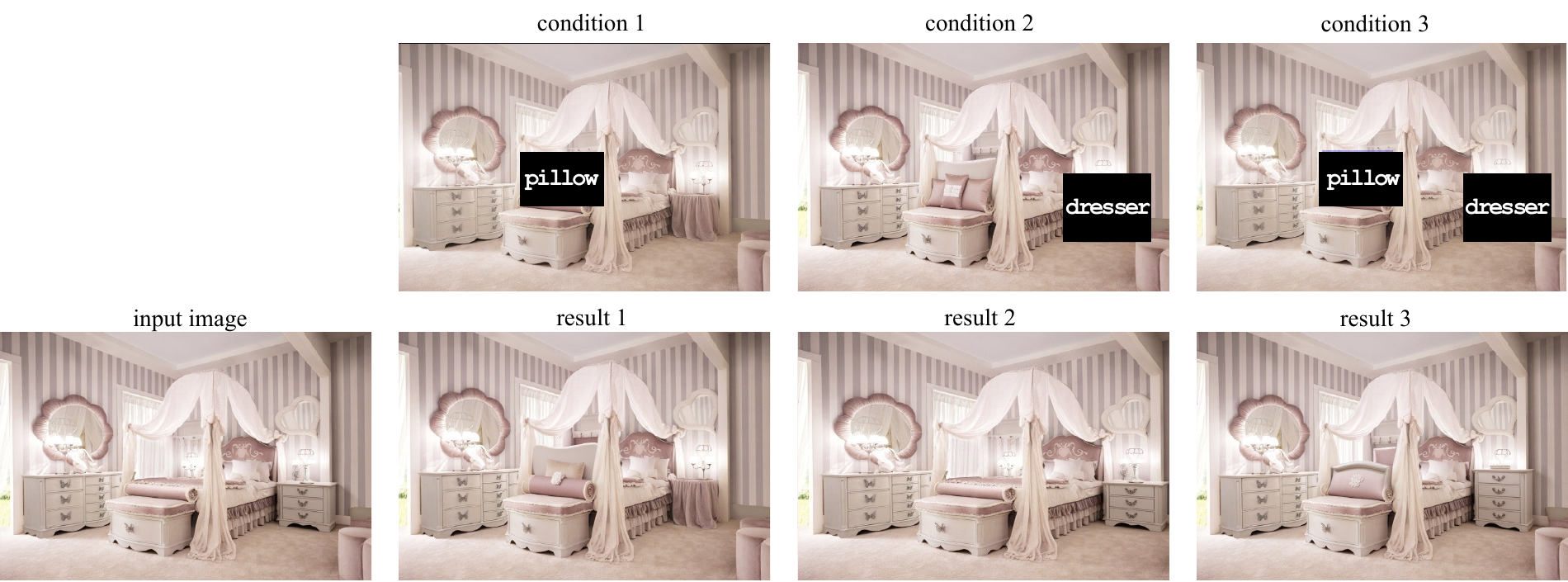}
    \caption{Multi-region edit modes. The source image is shown at the bottom-left. \emph{Columns 2--3 (multi-step edition):} the user applies a ``pillow'' mask (column~2) and then, on top of that result, a ``dresser'' mask (column~3); each step is a separate \method{} forward pass. \emph{Column 4 (single-step multibox edit):} both ``pillow'' and ``dresser'' masks are applied jointly in a single forward pass. \method{} handles both modes without architectural changes, producing spatially accurate edits while preserving the rest of the scene.}
    \label{fig:multibox}
\end{figure}

\method{} handles two complementary multi-region workflows in a single architecture: (i)~\emph{sequential multi-step edits}, where the user applies one mask at a time and the model's output from a previous step becomes the source image for the next; and (ii)~\emph{single-step multibox edits}, where multiple labeled masks are placed on the same image and processed jointly in one forward pass. Fig.\ref{fig:multibox} demonstrates both modes on the same source.

\subsection{Failure Cases}
\label{app:failure}

\begin{figure}[h]
    \centering
    \includegraphics[width=\linewidth]{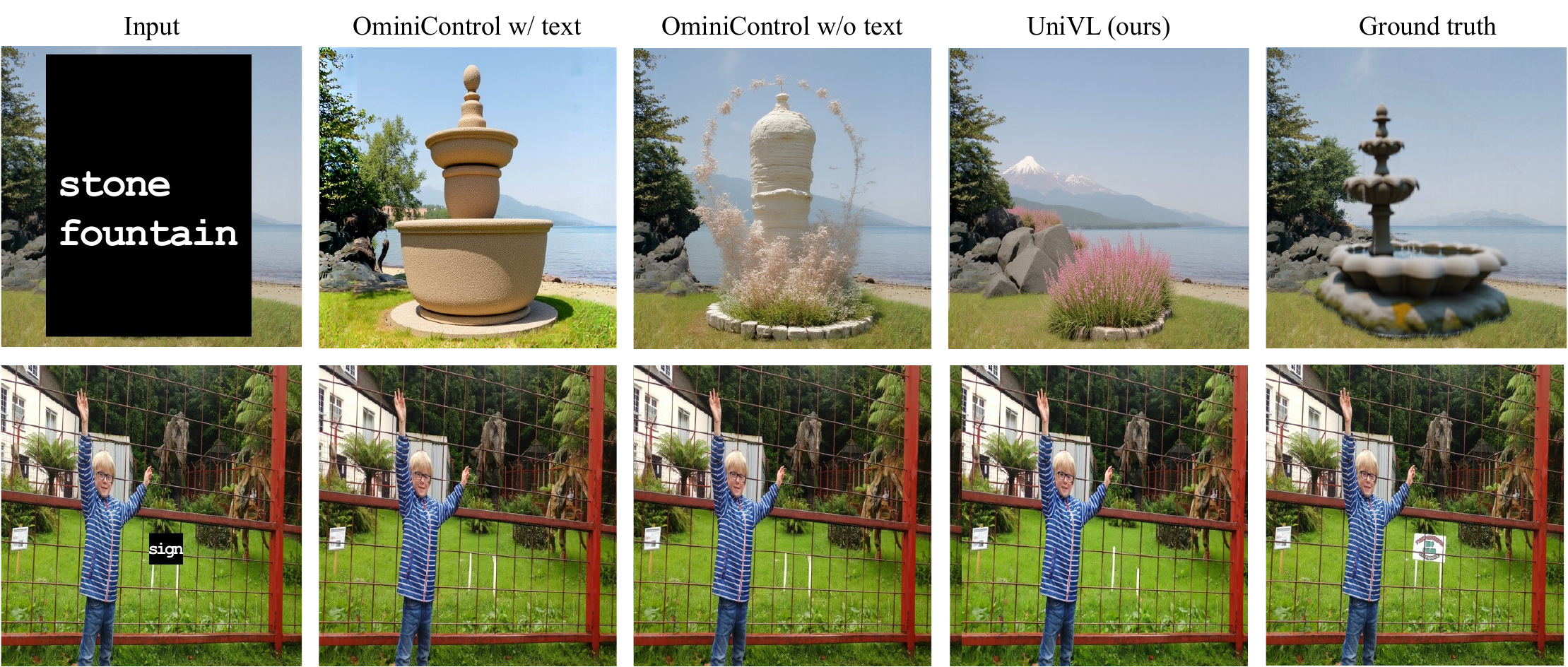}
    \caption{Failure cases of \method. \emph{(a)~Rare vocabulary:} for uncommon class phrases such as ``stone fountain,'' the rendered text is read correctly but the diffusion model has limited training signal for that category, producing generic stone-textured fills rather than the intended object. We expect this to improve with broader training-data coverage. \emph{(b)~Very small mask region:} when the mask area is tiny, the rendered text occupies only a few pixels and the \method{} encoder loses semantic resolution; the generated content drifts toward the background prior.}
    \label{fig:failure_cases}
\end{figure}

Fig.\ref{fig:failure_cases} illustrates two systematic failure modes.

Both failure modes share a common cause: insufficient signal for the \method{} encoder to extract a confident semantic instruction. Rare-vocabulary failures reflect a training-distribution gap that can be closed by augmenting the benchmark with longer-tail class phrases (the Summary section discusses this as future work); very-small-mask failures reflect a fundamental resolution limit on text-as-pixels conditioning, which scales with mask area.

\section{\method-ImgGen Benchmark Details}
\label{app:benchmark}

\subsection{Dataset categories and sources}
\label{app:benchmark_categories}

\method-ImgGen comprises four source categories that together provide both spatial-localization signal (mask category) and same-image-different-text signal (the three editing categories), preventing the model from collapsing to a pure image-reconstruction shortcut. Table~\ref{tab:benchmark_breakdown} summarizes the per-category statistics.

\begin{table}[h]
    \centering
    \caption{\method-ImgGen breakdown by source. The \emph{mask} category provides multibox spatial training; the three editing categories provide the same image with different text labels, ensuring the model uses the rendered text rather than reconstructing the image condition.}
    \label{tab:benchmark_breakdown}
    \begin{tabular}{lrrr}
        \toprule
        Category & Records (boxes) & Unique images & Boxes / image \\
        \midrule
        mask    & 280{,}900 & 23{,}375  & 12.02 \\
        add     & 122{,}594 & 122{,}594 & 1.00 \\
        replace & 217{,}606 & 217{,}606 & 1.00 \\
        extract & 131{,}996 & 131{,}996 & 1.00 \\
        \midrule
        Overall & 528{,}376 & 476{,}871 & 1.11 \\
        \bottomrule
    \end{tabular}
\end{table}

\paragraph{(1) \emph{mask} category (multibox spatial training).} Following the construction sketched in Section~\ref{sec:benchmark}, we start from a 238K-image subset of LAION-5B~\citep{schuhmann2022laion}, apply Grounding DINO~\citep{liu2023grounding} to detect objects and obtain bounding boxes with class labels, and apply two filters: (i)~a \emph{size filter} that discards detections whose mask area is less than 4\% of the total image area; and (ii)~a \emph{semantic filter} that retains pairs with CLIP similarity ${>}0.85$ between the cropped mask region and the class name. For each retained sample, the contextual condition is constructed by masking the detected region and rendering the class label as tiled text within it (Appendix~\ref{app:text_render}). The mask category has many boxes per image (12.02 on average), which we exploit through an \emph{augmentation} step: for each unique image, we sample 5 augmented duplicates with random crop and color jittering, each retaining 2--10 boxes. After augmentation, the mask category contains 233{,}750 final image instances, and the total number of records across the benchmark is 710{,}621.

\paragraph{(2--4) \emph{add}, \emph{replace}, and \emph{extract} categories (image editing).} These three categories are constructed following ImgEdit~\citep{ye2025imgedit}, with captions generated by GPT-4o and the corresponding image edits produced by GPT-4o-Image:
\begin{itemize}[leftmargin=1.5em]
    \item \textbf{Add} (122{,}594 pairs): inserting a new object into a designated region.
    \item \textbf{Replace} (217{,}606 pairs): editing the masked region with a new text-described object (text-prompt-driven replacement).
    \item \textbf{Extract} (131{,}996 pairs): an image-reference variant in which the masked region is filled by transferring content from a reference image; following ImgEdit~\citep{ye2025imgedit}, the class label of the rendered text in this category is the reference image's label.
\end{itemize}
Each editing-category record corresponds to a single image with one labeled mask, providing a complementary signal to the mask category: by training on many distinct text labels for the same source image (across the three editing categories), the model is encouraged to use the rendered text as a real conditioning signal rather than to reconstruct the image input.

\subsection{Class-name vocabulary statistics}
\label{app:benchmark_vocab}

\begin{figure}[t]
    \centering
    \begin{subfigure}[t]{0.4\linewidth}
        \centering
        \includegraphics[width=\linewidth]{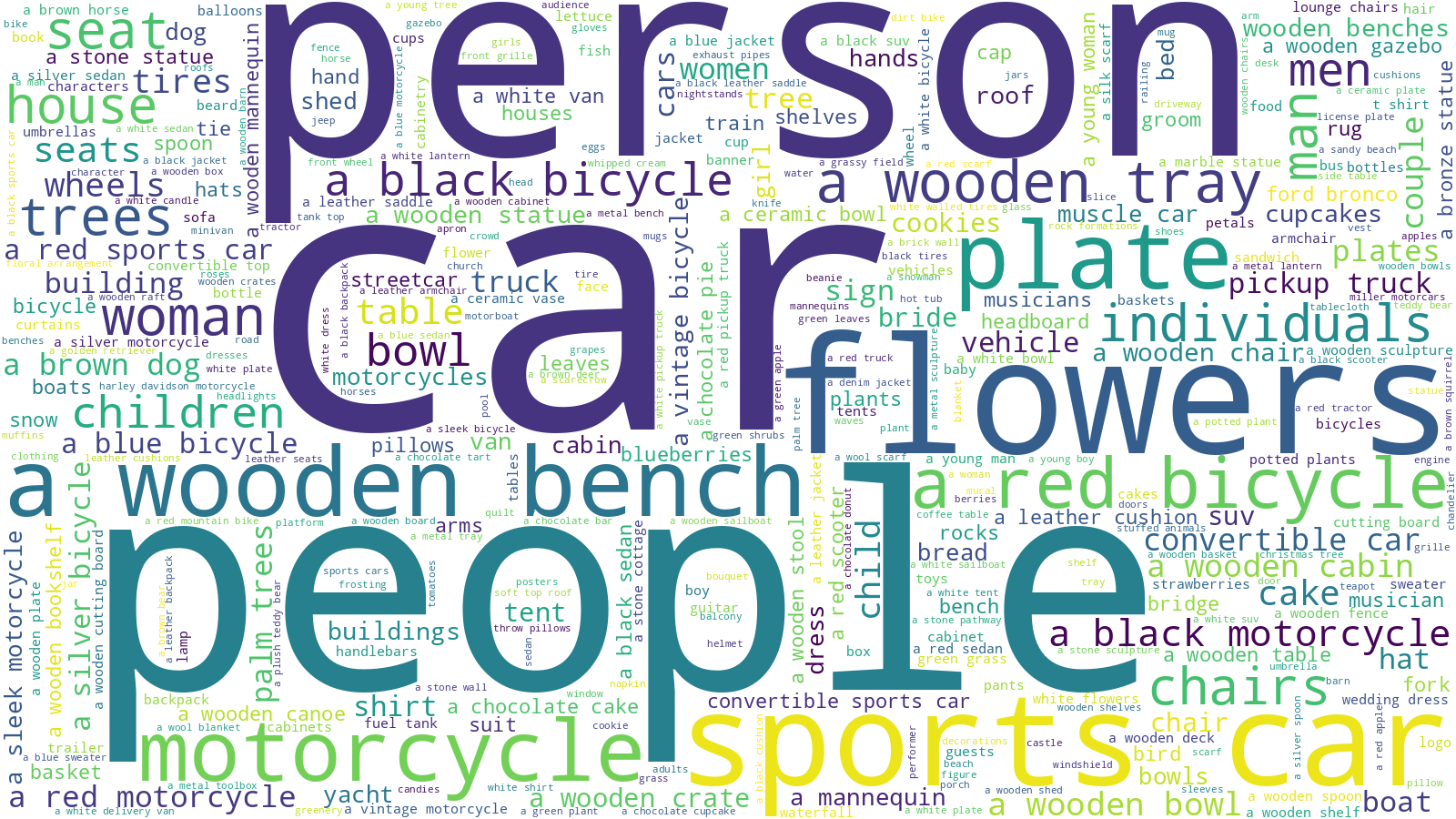}
        \caption{Word cloud over all class-name phrases.}
        \label{fig:vocab_wordcloud}
    \end{subfigure}
    \hfill
    \begin{subfigure}[t]{0.59\linewidth}
        \centering
        \includegraphics[width=\linewidth]{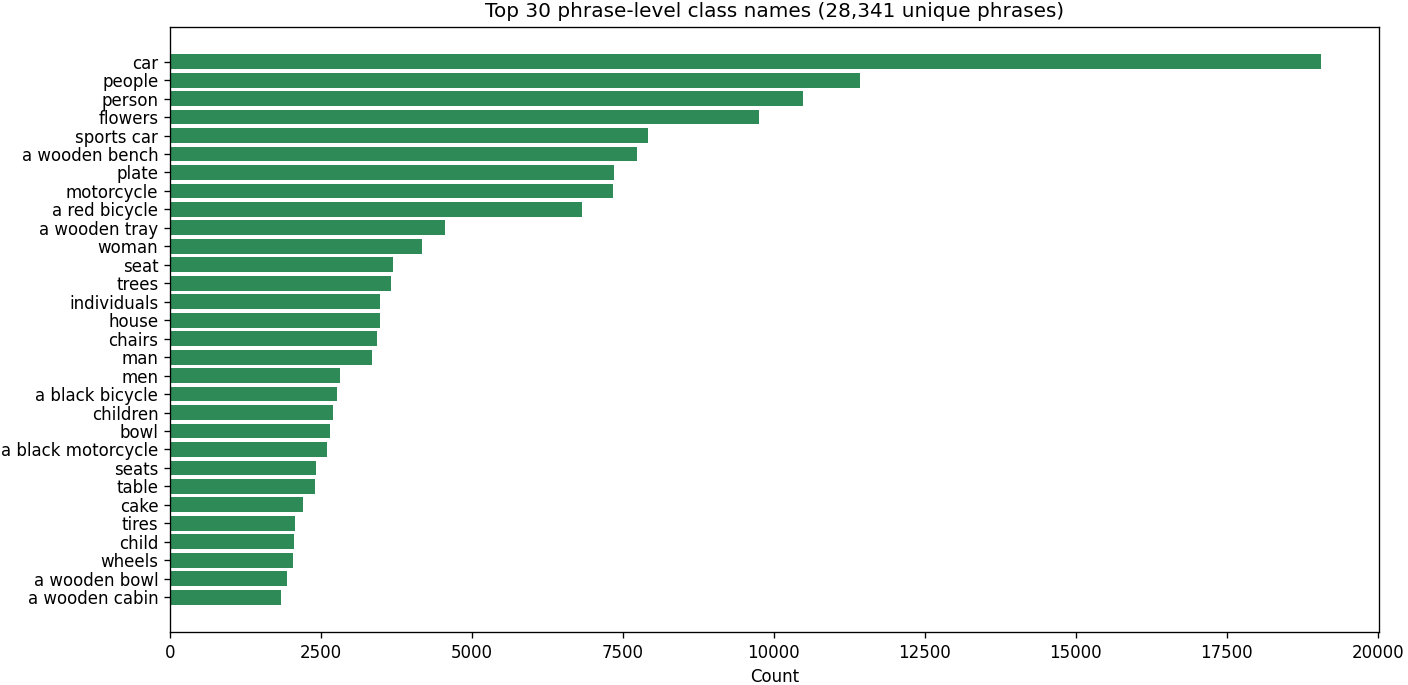}
        \vspace{-1em}
        \caption{Top-30 most frequent phrases.}
        \label{fig:vocab_top40_chart}
    \end{subfigure}
    \caption{Class-name vocabulary in \method-ImgGen. (\ref{fig:vocab_wordcloud})~Word cloud where font size is proportional to frequency: the most prominent labels are everyday objects and people (\emph{person}, \emph{car}, \emph{people}, \emph{flowers}, \emph{bench}, \emph{trees}), with a long tail of attribute-modified phrases (\emph{a wooden tray}, \emph{a black bicycle}, \emph{a red sports car}). (\ref{fig:vocab_top40_chart})~Frequency chart of the top-30 most frequent phrases. The vocabulary is dominated by natural-image object categories, validating our object-level scope.}
    \label{fig:vocab_top40}
\end{figure}

The class labels rendered onto masks span a long-tail vocabulary of natural-image object descriptions. Table~\ref{tab:phrase_length} reports the distribution of label phrase length (in words). The vast majority of labels are short (1--3 words: $96.0\%$ of records), but the benchmark also contains longer multi-word phrases (4 or more words: 4.0\%) that test the \method{} encoder's ability to read longer rendered text. Across all records there are ${\sim}28$K unique phrases. Fig.\ref{fig:vocab_top40} visualizes the top-30 most frequent class-name phrases.

\begin{table}[t]
    \centering
    \caption{Phrase-length distribution of class-name labels in \method-ImgGen. Most labels are short (1--3 words: 96.0\% of records) but the benchmark also includes longer multi-word phrases.}
    \label{tab:phrase_length}
    \begin{tabular}{rrrrr}
        \toprule
        \# words & Records & \% records & Unique phrases & \% unique \\
        \midrule
        1  & 270{,}607 & 51.21\% & 3{,}221  & 11.37\% \\
        2  &  89{,}675 & 16.97\% & 9{,}687  & 34.18\% \\
        3  & 146{,}959 & 27.81\% & 10{,}778 & 38.03\% \\
        4  &  19{,}486 &  3.69\% & 3{,}937  & 13.89\% \\
        5  &   1{,}253 &  0.24\% &    536   &  1.89\% \\
        6  &       327 &  0.06\% &    149   &  0.53\% \\
        7  &        47 &  0.01\% &     23   &  0.08\% \\
        8  &        13 &  0.00\% &      7   &  0.02\% \\
        9  &         6 &  0.00\% &      2   &  0.01\% \\
        11 &         3 &  0.00\% &      1   &  0.00\% \\
        \bottomrule
    \end{tabular}
\end{table}

\subsection{Training and evaluation splits}
\label{app:benchmark_splits}

The 710K records form the training pool. To support multi-region training, 20\% of training batches use multibox data drawn from the mask category (2--5 non-overlapping masks per image; mask sampling rules in Appendix~\ref{app:multibox}). For evaluation, we hold out 3{,}000 samples, evenly split into:
\begin{itemize}[leftmargin=1.5em]
    \item \textbf{1{,}500 single-mask samples}: 375 per task type (mask, object addition, replacement, extraction).
    \item \textbf{1{,}500 multi-mask samples}: $N{\in}[1,5]$ non-overlapping masks per image, testing composable multi-region generation.
\end{itemize}
Metrics in Table~\ref{tab:main_results} are computed on the union of these 3{,}000 samples; per-mask-count breakdowns appear in Fig.\ref{fig:compression}(b)(c).

\end{document}